\documentclass[letterpaper]{article} 
\usepackage[draft]{aaai2026}  
\usepackage{times}  
\usepackage[table]{xcolor}
\usepackage{helvet}  
\usepackage{courier}  
\usepackage[hyphens]{url}  
\usepackage{graphicx} 
\usepackage{natbib}  
\usepackage{caption} 
\usepackage{algorithm}
\usepackage{algorithmic}

\usepackage{newfloat}
\usepackage{listings}
\DeclareCaptionStyle{ruled}{labelfont=normalfont,labelsep=colon,strut=off} 
\floatstyle{ruled}
\newfloat{listing}{tb}{lst}{}
\floatname{listing}{Listing}
\newcommand{\ignore}[1]{}

\usepackage{enumitem}
\usepackage{adjustbox}
\usepackage{booktabs}
\usepackage{multirow}
\usepackage{amsmath}
\usepackage{array}
\newcolumntype{H}{>{\setbox0=\hbox\bgroup}c<{\egroup}@{}}

\usepackage{arydshln}

\title{Evaluating, Synthesizing, and Enhancing for Customer Support Conversation}
\author {
    Jie Zhu\textsuperscript{\rm 1,2}\thanks{These authors contributed equally},
    Huaixia Dou\textsuperscript{\rm 1,2}\footnotemark[1],
    Junhui Li\textsuperscript{\rm 1}\thanks{Corresponding Author},
    Lifan Guo\textsuperscript{\rm 2}, \\
    Feng Chen\textsuperscript{\rm 2},
    Chi Zhang\textsuperscript{\rm 2},
    Fang Kong\textsuperscript{\rm 1}
}
\affiliations {
    \textsuperscript{\rm 1}School of Computer Science and Technology, Soochow University\\
    \textsuperscript{\rm 2}Qwen DianJin Team, Alibaba Cloud Computing\\
}

\begin{document}

\maketitle

\begin{abstract}
Effective customer support requires not only accurate problem-solving but also structured and empathetic communication aligned with professional standards. However, existing dialogue datasets often lack strategic guidance, and real-world service data is difficult to access and annotate. To address this, we introduce the task of Customer Support Conversation (CSC), aimed at training customer service supporters to respond using well-defined support strategies. We propose a structured CSC framework grounded in COPC guidelines, defining five conversational stages and twelve strategies to guide high-quality interactions. Based on this, we construct {\tt CSConv}, an evaluation dataset of 1,855 real-world customer–agent conversations rewritten using LLMs to reflect deliberate strategy use, and annotated accordingly. Additionally, we develop a role-playing approach that simulates strategy-rich conversations using LLM-powered roles aligned with the CSC framework, resulting in the training dataset {\tt RoleCS}. Experiments show that fine-tuning strong LLMs on {\tt RoleCS} significantly improves their ability to generate high-quality, strategy-aligned responses on {\tt CSConv}. Human evaluations further confirm gains in problem resolution. All code and data will be publicly available at \url{https://github.com/aliyun/qwen-dianjin}.

\end{abstract}


\section{Introduction}\label{sec:introduction}

\begin{figure}[!ht]
\centering
\includegraphics[width=0.9\linewidth]{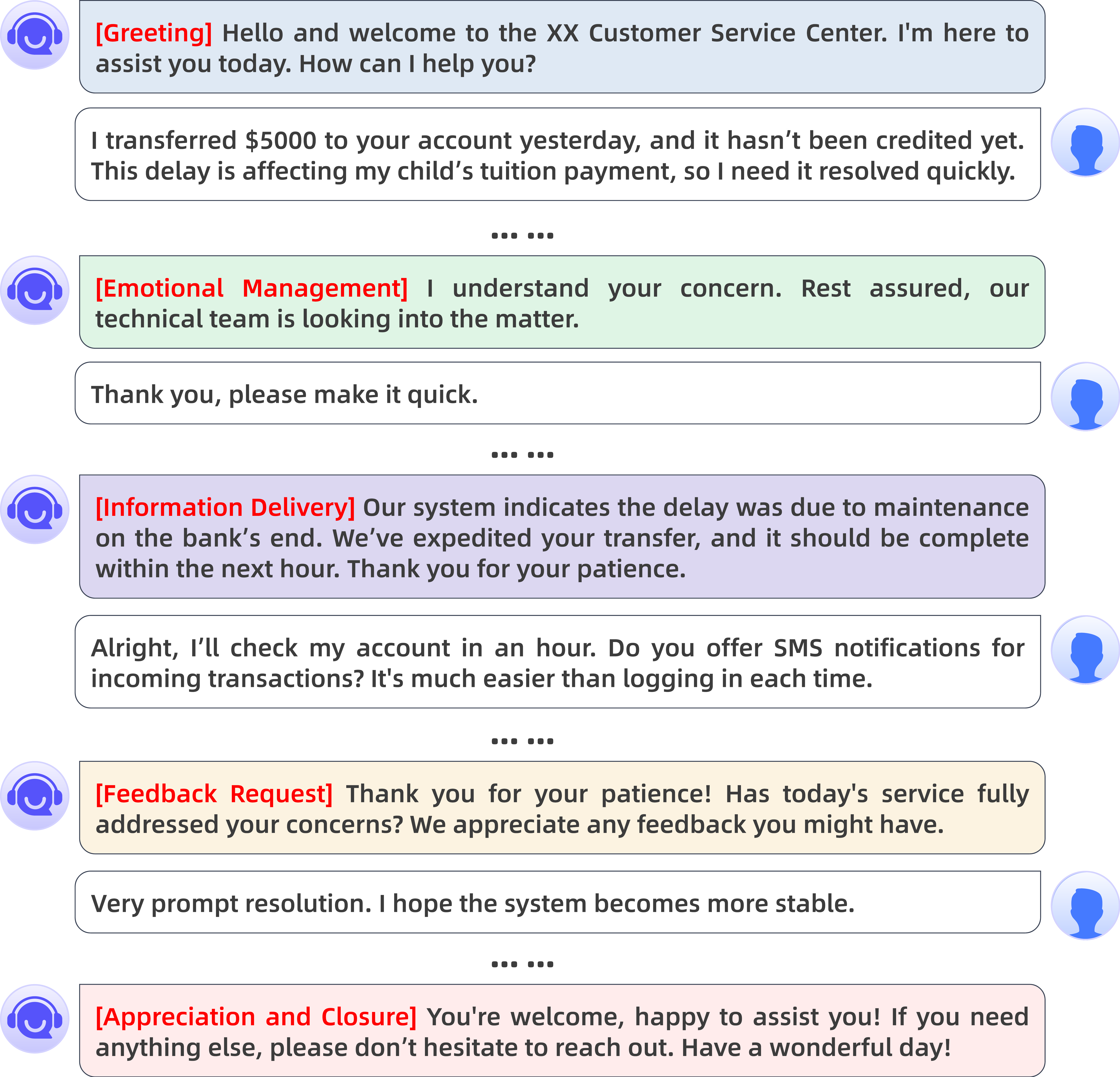}
\caption{An example dialogue between a service supporter (left) and a customer (right), showing \textbf{support strategies} (noted in parentheses) used by the supporter. The conversation is organized into five stages of the proposed CSC framework, shown in colored boxes.}
\label{fig:intro_example}
\end{figure}

Customer support aims to help users resolve product- or service-related issues through effective and context-aware communication. While large-scale dialogue systems have received growing attention in recent years~\cite{budzianowski-etal-2018-multiwoz,rashkin-etal-2019-towards,peskov-etal-2019-multi,liu-etal-2021-towards}, customer support conversations remain underexplored in the NLP community,\footnote{This paper uses \textit{dialogue} and \textit{conversation} interchangeably.} largely due to the scarcity of publicly available benchmarks and the sensitive, domain-specific nature of support interactions. Meanwhile, large language models (LLMs) like GPT-3~\cite{brown-etal-2020-language} have shown impressive capabilities in open-domain dialogue generation~\cite{zhang-etal-2020-dialogpt,bae-etal-2022-building,thoppilan-etal-2022-lamda}, but their ability to generate realistic and effective customer support conversations remains underexamined.

Customer support guidelines from COPC, the internationally recognized standard for customer experience management,\footnote{https://www.copc.com/copc-standards/} emphasize that high-quality support often depends on structured communication strategies, similar to those used in emotional support settings~\cite{liu-etal-2021-towards}. As shown in Figure~\ref{fig:intro_example},\footnote{The example is translated into English for better readability.} the service supporter begins with the \textit{Greeting} strategy to initiate the conversation and explore the customer's issue. Upon recognizing the customer's negative emotion, the supporter adopts the \textit{Emotional Management} strategy, expressing empathy and understanding to alleviate the customer's distress. After understanding the problem, the supporter employs the \textit{Information Delivery} strategy to provide clear and actionable guidance, followed by the \textit{Feedback Request} strategy to check for further concerns. Finally, the conversation ends with the \textit{Appreciation and Closure} strategy, ensuring a respectful and reassuring followed by \textit{Feedback Request} to check for further concerns.

Despite the importance of effective customer support, research on real-time customer service conversations remains limited, mainly due to a lack of task-specific design and high-quality annotated data. Many studies~\cite{xu-etal-2017-a,oraby-etal-2017-how,cui-etal-2017-superagent,mesquita-etal-2022-dense} rely on asynchronous, exchanges (e.g., Twitter), where interactions span minutes to days. These settings differ significantly from the immediate, uninterrupted nature of real-time support. Moreover, effective support requires not only resolving issues but also showing empathy and emotional support. Yet, most task-oriented dialogue datasets~\cite{wen-etal-2017-network,budzianowski-etal-2018-multiwoz,peskov-etal-2019-multi,gung-etal-2023-natcs} lack the intentional use of supportive strategies like emotional management or empathetic closure, which are vital for high-quality customer service.

To address this gap, we introduce the task of {\bf c}ustomer {\bf s}upport {\bf c}onversation (CSC), to facilitate the training of customer service supporters. The goal is to help them respond with appropriate strategies that combine accurate solutions with empathetic communication. Based on COPC standards and inspired by the emotional support framework~\cite{liu-etal-2021-towards}, we develop a CSC-specific framework with five dialogue stages and twelve support strategies. Using it, we build \texttt{CSConv}, a high-quality dataset adapted from real service interactions and refined for structured strategy use. To address the lack of high-quality training data, we also develop a role-playing approach that creates \texttt{RoleCS}, a synthetic dataset of strategy-rich conversations by assigning LLMs distinct roles aligned with the CSC framework.\ignore{Moreover, to address the lack of high-quality training data, we develop a role-playing approach that simulates strategy-rich conversations by assigning LLMs distinct roles aligned with the CSC framework, resulting in the synthetic dataset {\tt RoleCS}.} Fine-tuning on {\tt RoleCS} significantly boosts LLMs' ability to generate strategy-aligned and effective responses on {\tt CSConv}.

\section{Related Work}\label{sec:related_work}

\subsection{Task-Oriented Conversation Datasets.}
The goal of {\texttt{CSConv}}, which is to generate conversations based on real spoken customer service interactions, differs significantly from previous task-oriented dialogue datasets. Many earlier synthetic datasets have been created using the Wizard of Oz (WOZ) framework~\cite{kelley-1984-an}, where one person acts as the system and another as the user. \citet{wen-etal-2017-network} introduce a crowdsourced version of the WOZ framework to collect domain-specific dialogue data more efficiently. Other task-oriented dialogue datasets include Frames~\cite{asri-etal-2017-frames}, MultiWOZ~\cite{budzianowski-etal-2018-multiwoz}, MultiDoGO~\cite{peskov-etal-2019-multi}, EMPATHETICDIALOGUES~\cite{rashkin-etal-2019-towards}, Taskmaster-1~\cite{byrne-etal-2019-taskmaster}, ESConv~\cite{liu-etal-2021-towards}, and NATCS~\cite{gung-etal-2023-natcs}, each of which explores different domains and annotation schemes for developing conversational agents. 

Among them, ESConv is closely related to our work, as both it and {\tt CSConv} aim to enhance service quality using structured support strategies. However, their construction methods differ significantly: ESConv uses a crowdsourced WOZ setup, while {\tt CSConv} rewrites real-world service dialogues using high-performing LLMs.

\subsection{Role-Playing in Conversation Generation}
Recent advances in LLMs have enabled the use of role-playing agents for conversation generation. For example, \citet{bae-etal-2022-building} generate dialogues aligned with specific roles, while \citet{yang-etal-2024-crafting} simulate characters and settings to produce coherent interactions. \citet{wu-etal-2024-from} create interactive narrative dramas, and \citet{ye-etal-2025-sweetiechat} focus on emotional support conversations. Building on the \citet{ye-etal-2025-sweetiechat}, we design role-playing agents for customer support with clearer role definitions and responsibilities to more closely reflect real-world service interactions. 

\begin{figure*}[t]
\centering
\includegraphics[width=0.8\textwidth]{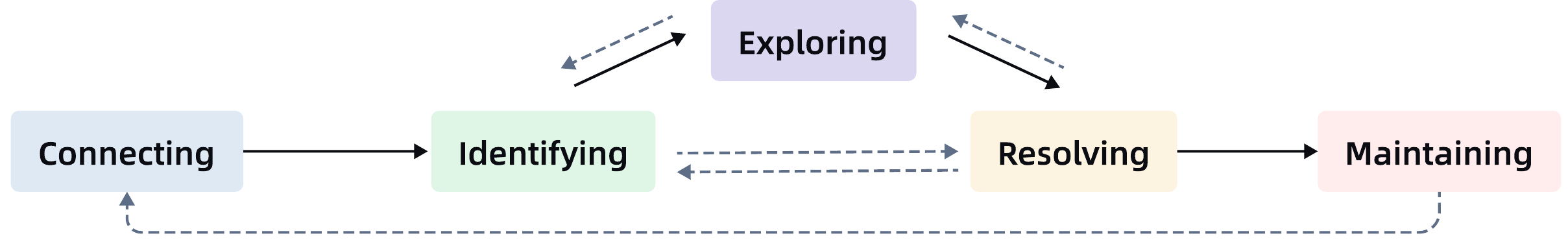}
\caption{Overview of the CSC framework's five stages, each paired with recommended support strategies (see Table~\ref{tbl:strategy} in Appendix~\ref{apdx:csc_framework}). The typical flow is: \textcircled{1} Connecting → \textcircled{2} Identifying → \textcircled{3} Exploring → \textcircled{4} Resolving → \textcircled{5} Maintaining (black arrows), but it can be adjusted based on the specifics of each conversation (dashed arrows).}
\label{fig:stage_flow}
\end{figure*}

\section{CSC: Customer Support Conversation}\label{sec:csc}
In customer support conversation (CSC), the service supporter aims to accurately identify customer issues while also addressing emotional needs. This dual focus enables effective solutions that enhance customer satisfaction and improve problem-resolution efficiency. The supporter is considered effective when both the issue and emotional concerns are properly addressed. To build an evaluation dataset for CSC, we first introduce the CSC framework in Section~\ref{sec:csc_framework}, followed by the data construction process in Section~\ref{sec:data_construction} and dataset statistics in Section~\ref{sec:dataset_analysis}. Finally, we define the CSC task in Section~\ref{sec:task_define}.

\subsection{CSC Framework}\label{sec:csc_framework}

The CSC framework organizes the customer support process into five stages, each with recommended strategies designed to enhance the quality and effectiveness of interaction.

\paragraph{Stages.} Building on the three core stages of supportive communication from \citet{hill-2019-help} and the COPC practical guidelines, we work with domain experts to refine and extend this structure for customer support. The resulting CSC framework defines five stages: \textit{Connecting} (greeting and rapport building), \textit{Identifying} (understanding the customer's issue and emotional state), \textit{Exploring} (discussing and evaluating potential solutions), \textit{Resolving} (delivering and confirming resolution), and \textit{Maintaining} (closing the interaction while preserving the customer relationship). Figure~\ref{fig:stage_flow} illustrates the flow across these stages.

Importantly, these stages are not rigid steps but modular components that can appear in various orders or combinations depending on the nature of the conversation. For example, even when no solution is reached, a supporter may still provide empathy (\textit{Exploring}), acknowledge service limitations (\textit{Resolving}), and close the conversation professionally (\textit{Maintaining}). This flexible structure enables consistent analysis and strategy modeling across both successful and challenging customer support scenarios.

\paragraph{Strategies.} In parallel, we work with domain experts to define twelve actionable support strategies aligned with the five stages: \textit{Greeting (GT)}, \textit{Identity Verification (IV)}, \textit{Emotional Management (EM)}, \textit{Restatement or Paraphrasing (RP)}, \textit{Problem Refinement (PR)}, \textit{Providing Suggestions (PS)}, \textit{Information Delivery (ID)}, \textit{Resolution Implementation (RI)}, \textit{Feedback Request (FR)}, \textit{Appreciation and Closure (AC)}, \textit{Relationship Continuation (RC)}, and \textit{Others}. These strategies provide practical guidance for managing both task-related issues and emotional support throughout the customer interaction. For the details of stages and strategies, please refer to Appendix~\ref{apdx:csc_framework}.

\subsection{Dataset Construction}\label{sec:data_construction}

We construct the {\tt CSConv} dataset using Chinese customer service dialogues collected from \ignore{real-world service scenarios provided by }our pre-sales and after-sales customer service centers. Prior to our access, all conversations were professionally transcribed, manually corrected, and fully de-identified to ensure privacy protection and integrity. While these raw conversations authentically reflect real-world customer interactions, they often lack consistent structure, making it difficult to systematically annotate support strategies according to the COPC-informed CSC framework. To address this, we employ an LLM to rewrite the conversations while preserving the original semantics and user intent. This controlled rewriting aligns conversations with the defined dialogue stages and strategies, enabling more consistent and interpretable annotations without sacrificing the complexity of real user queries. 

\textit{Importantly, the goal of constructing \texttt{CSConv} is not to evaluate real-time chatbot performance, but to facilitate the training of customer service supporters by helping them learn to respond using appropriate strategies guided by the COPC framework.} In total, we collect 690K conversations spanning eight in-domain topics. To ensure quality, we use a four-stage pipeline to guide data selection and refinement.

\begin{enumerate}[leftmargin=*]
\item Pre-filtering: We first apply rule-based filtering to remove low-quality conversations. These rules exclude conversations that are too short or too long, contain overly lengthy utterances, show an imbalance between customer and agent turns, or have a high proportion of ineffective customer responses. Additionally, we use an LLM to exclude conversations with offensive or unprofessional content.

\item Sampling and Rewriting: Up to 500 filtered conversations per topic are sampled and rewritten by an LLM to align with the CSC framework. During this process, the LLM analyzes the original scenario and generates a new conversation that preserves the core issue while improving clarity, structure, and emotional engagement. For each agent turn, the LLM selects an appropriate support strategy based on conversation context, occasionally using \textit{Others} to maintain conversational naturalness. Customer responses are also refined for coherent interaction. 

\item Post-filtering: After rewriting, a second round of rule-based and LLM-based checks verifies structure (e.g., strategy coverage, speaker alternation, and length) and filters out conversations lacking coherence, empathy, or strategic alignment. 

\item Manually annotation: Finally, experts certified in COPC review the remaining conversations, evaluating the support agent's responses for realism, empathy, and adherence to the CSC framework. This results in a curated evaluation set of 1,855 high-quality conversations. 
\end{enumerate}

Appendix~\ref{apdx:more_data_construction} provides details on filtering rules, prompt templates, and annotation guidelines. For rewriting, we use \texttt{DeepSeek-R1}~\cite{guo-etal-2025-deepseek-r1} instead of \texttt{GPT-4o}~\cite{openai-2024-gpt4o}, as the latter tends to produce shorter, less emotionally rich dialogues. See Appendix~\ref{apdx:model_comparison} for a comparison.

\subsection{Dataset Analysis}\label{sec:dataset_analysis}

\paragraph{Overview Statistics.} Table~\ref{tbl:testset} compares statistics of {\tt CSConv} before and after rewriting, covering 1,855 conversations. Rewritten dialogues are longer, averaging 27.27 versus 19.06 utterances, with supporter responses increasing from 41.16 to 48.72 words and customer responses decreasing from 21.60 to 17.17. Notably, strategy use (excluding \textit{Others}) rises from 55.28\% to 97.82\%,\footnote{We use \texttt{Qwen2.5-72B-Instruct} to assign a support strategy to each supporter response in the dialogues prior to rewriting.} indicating more deliberate and guided support. These shifts reflect the goal of rewriting, which is to facilitate the training of customer service supporters in applying appropriate strategies aligned with the COPC framework.\ignore{This shift suggests a more proactive and informative role by the supporter, who provides clearer and more empathetic guidance, while helping customers articulate their concerns more succinctly. }

\paragraph{Topic Distribution.} Table~\ref{tbl:scenarios} presents the distribution of conversations across eight topic. Each topic, excluding \textit{Others}, accounts for roughly 11\% to 16\% of the dataset.

\begin{table}[t]
\centering
\small
\begin{adjustbox}{width=\columnwidth}
\begin{tabular}{llll}
\toprule[1pt]
\multirow{2}{*}{-}& \multirow{2}{*}{-} & \multicolumn{2}{c}{\textbf{Number}} \\
& & \textbf{Original} & \textbf{Rewritten} \\
\hline
\multirow{4}{*}{Total} & Conversations & 1,855 & 1,855 \\
& Utterances & 35,350 & 50,587 \\
& Avg. Utterance Number & 19.06 & 27.27 \\
& Avg. Utterance Length	& 31.48 & 33.27 \\
\hline
\multirow{3}{*}{Supporter} & Utterances & 17,862 & 25,810 \\
& Avg. Utterance Number & 9.63 & 13.91 \\
& Avg. Utterance Length	& 41.16 & 48.72 \\
& Strategy Use (w/o \textit{Others}) & 55.28\% & 97.82\% \\
\hline
\multirow{3}{*}{Customer} & Utterances & 17,488 & 24,777 \\
& Avg. Utterance Number & 9.43 & 13.36 \\
& Avg. Utterance Length	& 21.60 & 17.17 \\
\bottomrule[1pt]
\end{tabular}
\end{adjustbox}
\caption{Statistics of {\tt CSConv} before and after rewriting.}
\label{tbl:testset}
\end{table}

\begin{table}[t]
\centering
\begin{adjustbox}{width=\columnwidth}
\begin{tabular}{lll}
\toprule[1pt]
\textbf{Topic} & \textbf{Num} & \textbf{Proportion} \\
\hline
Account and Transaction Management & 265 & 14.3\% \\
Product Consultation & 242 & 13.0\% \\
Technical Support and Online Services & 295 & 15.9\% \\
Complaints and Dispute Resolution & 263 & 14.2\% \\
Marketing and Promotion Activities & 211 & 11.4\% \\
Risk Management and Security & 266 & 14.3\% \\
Financial Consulting and Planning & 263 & 14.2\% \\
Others & 50 & 2.7\% \\
\hline
Overall & 1,855 & 100.0\% \\
\bottomrule[1pt]
\end{tabular}
\end{adjustbox}
\caption{Distribution of topics in {\tt CSConv}.}
\label{tbl:scenarios}
\end{table}


\begin{figure}[t]
\centering
\includegraphics[width=\columnwidth]{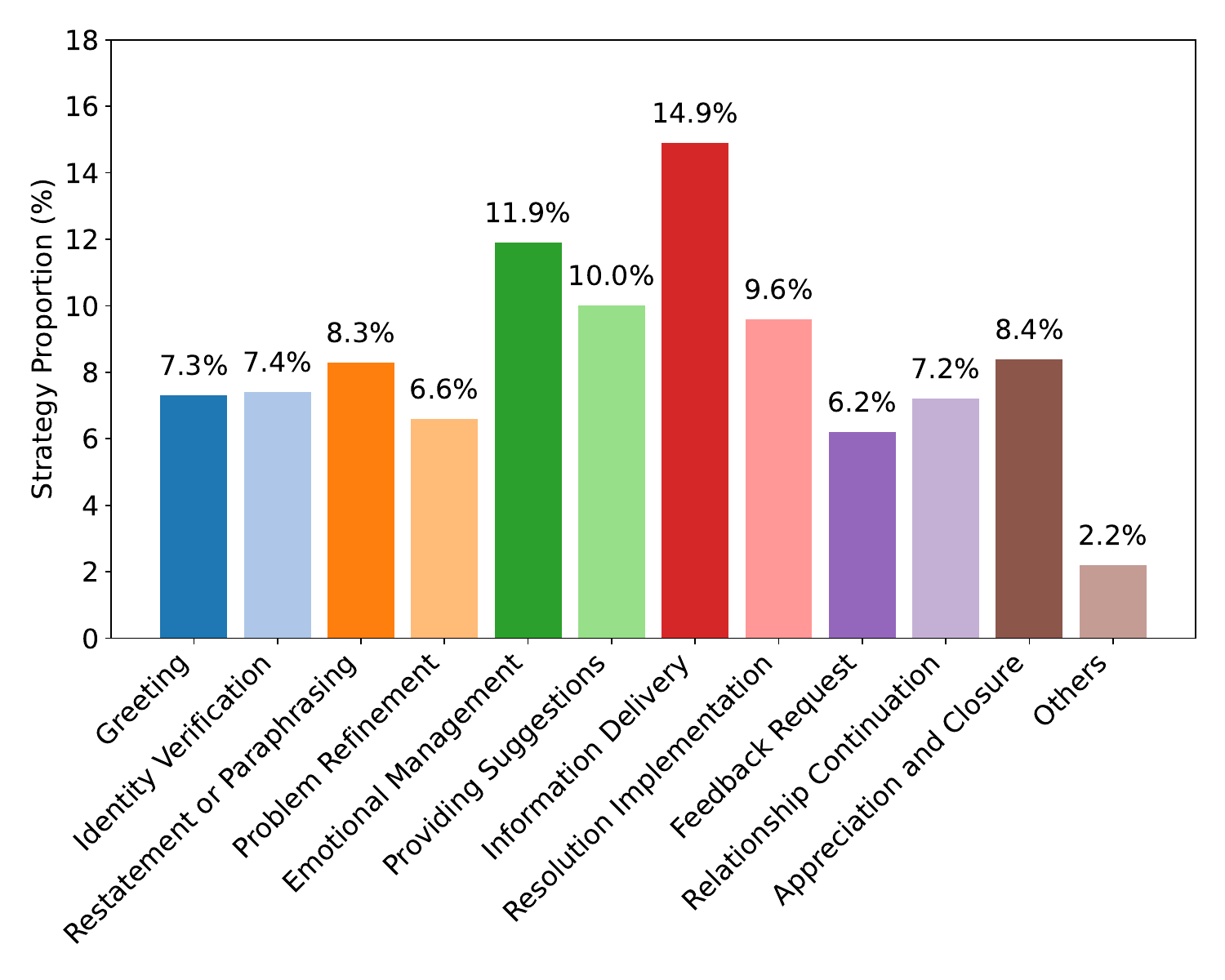}
\caption{Strategy proportion of {\tt CSConv}.}
\label{fig:strategy_proportion_testset}
\end{figure}

\paragraph{Strategy Distribution.} As shown in Figure \ref{fig:strategy_proportion_testset}, the most commonly used strategies are \textit{Information Delivery} (14.9\%), \textit{Emotional Management} (11.9\%), and \textit{Provide Suggestions} (10.0\%). This highlights the dual importance of delivering clear information and managing customer emotions in effective support conversations.

Additional statistics on strategy transitions and unique strategy distributions are provided in Appendix~\ref{apdx:more_statistics}.

\subsection{CSC Task Definition}\label{sec:task_define}

We denote a customer support conversation as {\small $D = \{\left(P_i, T_i, U_i\right)\}_{i=1}^N$}, consisting of {\small $N$} turns exchanged between a supporter {\small $S$} and a customer {\small $C$}. For each turn, {\small $P_i\in \{S, C\}$} denotes the speaker, {\small $U_i$} is the utterance text, and {\small $T_i$} represents the response strategy used. Strategies are selected from a predefined set {\small $G$}, and are assigned only to the supporter's turns. That is, if {\small $P_i = C$}, then {\small $T_i = \text{NULL}$}.

Following the task of emotional support conversation~\cite{liu-etal-2021-towards,ye-etal-2025-sweetiechat}, we define the CSC task as generating the supporter's response. Specifically, at turn {\small $k$}, where {\small $P_k=S$}, the model receives the conversation history {\small $X_k=\{\left(P_i, T_i, U_i\right)\}_{i=1}^{k-1}$} as input. The CSC task is then divided into two subtasks:\footnote{We further investigate how strategy prediction contributes to improving response generation in Section~\ref{sec:effect_strategy}.}
\begin{enumerate}[leftmargin=*]
\item \textbf{Strategy Prediction:} Predict the appropriate support strategy {\small $T_k \in G$} based on the conversation history {\small $X_k$}. 
\item \textbf{Response Generation:} Generate a response {\small $U_k$} conditioned on both the predicted strategy {\small $T_k$} and the conversation history {\small $X_k$}, ensuring that the output is consistent with both the customer's needs and the strategic intent. 
\end{enumerate}

\section{Synthetic Conversation Generation with Role-Playing Agents}
\begin{figure*}[!ht]
    \centering
    \includegraphics[width=\textwidth]{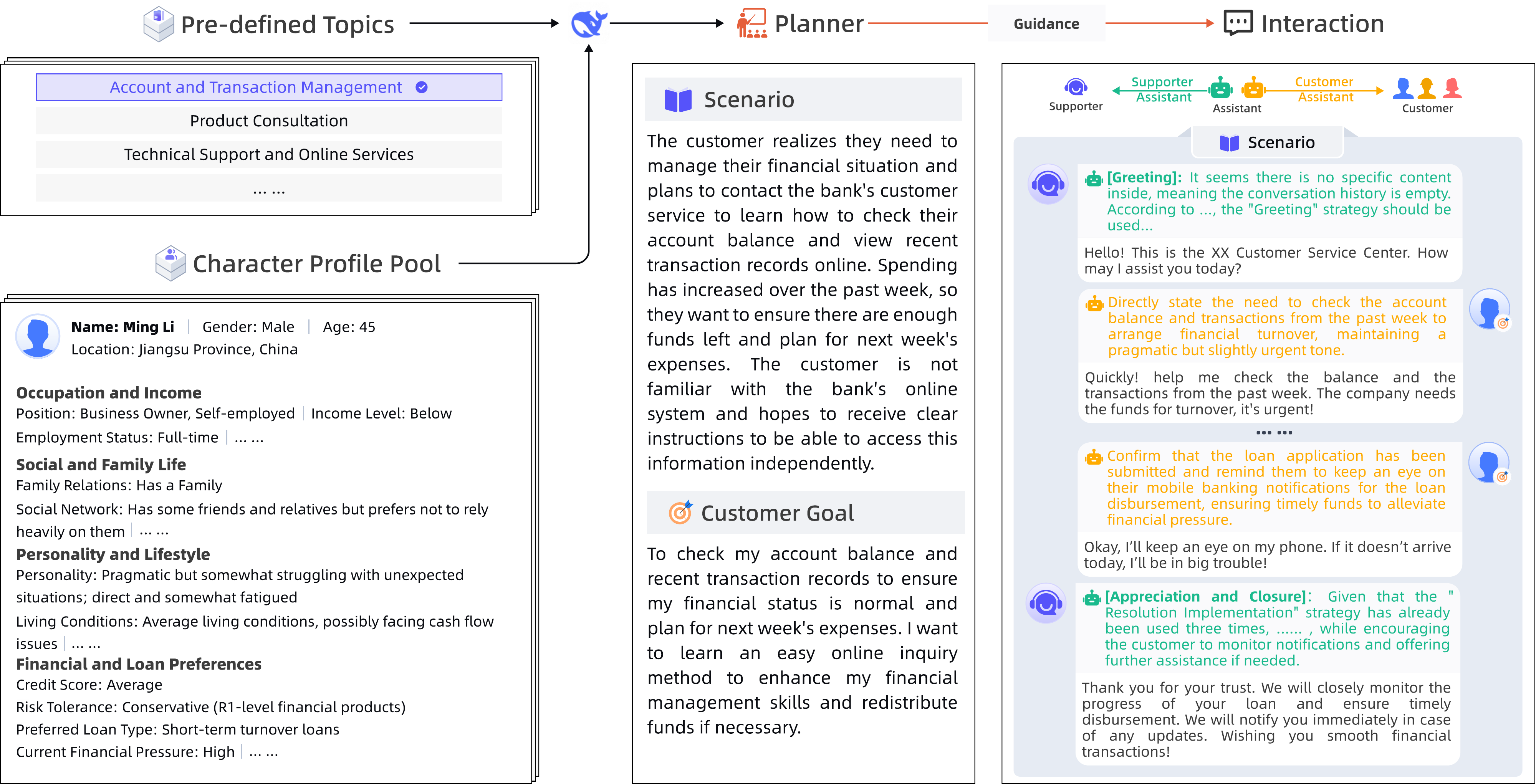}
    \caption{Illustration of synthetic conversation generation using role-playing agents.}
    \label{fig:training}
\end{figure*}

Fine-tuning LLMs has shown effective for improving performance in task-specific dialogue applications. However, building high-quality, multi-turn customer support datasets remains challenging due to the need for domain expertise, contextual coherence, and dialogue diversity. This limits both the scalability and scenario coverage. While recent studies use LLMs for dataset augmentation via rewriting or imitation, they often yield limited variation in dialogue flow and support strategy. To address this, inspired by \citet{ye-etal-2025-sweetiechat}, we adopt a multi-role role-playing framework to generate synthetic customer support dialogues that are diverse, coherent, and representative of real-world scenarios.

\subsection{Role Playing Conversation Generation}
\label{sec:role_construction}

As shown in Figure \ref{fig:training}, our role-playing framework involves five roles: \textit{Planner}, \textit{Supporter Assistant}, \textit{Supporter}, \textit{Customer Assistant}, and \textit{Customer}. Each role serves a distinct purpose. The \textit{Planner} defines the dialogue scenario and sets the \textit{Customer}'s communication goal. The main interaction occurs between the \textit{Supporter} and the \textit{Customer}, simulating a real customer service exchange. Meanwhile, their respective \textit{Assistants}, the \textit{Supporter Assistant} and the \textit{Customer Assistant}, offer strategic guidance to help them fulfill their roles more effectively. We use \texttt{deepseek-r1}~\cite{guo-etal-2025-deepseek-r1} to simulate all roles, with detailed prompts available in Appendix \ref{apdx:prompts_for_agent}. We describe each role below.

\subsubsection{Planner}\label{sec:planner}
Before the conversation begins, the Planner selects a topic {$e$} from a predefined list {$E$} of customer topics in Table~\ref{tbl:scenarios} (excluding \textit{Others}). It then samples a customer profile {$o$} from a character pool {$O$} that contains diverse customer personas. Using the selected topic {$e$} and customer profile {$o$}, the Planner prompts an LLM {$\mathcal{M}$} to generate a detailed service scenario {$e'$} along with a corresponding communication goal {$g$}, denoted as {$(g, e')=\mathcal{M}\left(o, e\right)$}. This setup ensures that each dialogue is grounded in a realistic context.

\subsubsection{Supporter Assistant}\label{sec:supporter_assistant} 
The Supporter Assistant recommends an appropriate support strategy {$t$} from the predefined strategy set {$G$}, based on the current dialogue history {$h_s$} and the scenario $e'$. The recommendation is generated by querying the LLM {$\mathcal{M}$}, i.e., {$t = \mathcal{M}(h_s, G, e')$}. This strategic guidance helps the Supporter stay aligned with the customer's needs and maintain coherent throughout the conversation.

\subsubsection{Supporter}\label{sec:supporter}
Guided by the support strategy {$t$} suggested by the Supporter Assistant, the Supporter generates a contextually appropriate response {$r_s$}. This response is produced by the LLM {$\mathcal{M}$}, conditioned on the dialogue history {$h_s$}, the strategy {$t$} and the scenario $e'$, i.e., {$r_s = \mathcal{M}(h_s, t, e')$}.

\subsubsection{Customer Assistant}\label{sec:customer_assistant} 
The Customer Assistant guides the conversation by generating the next direction {$d$}, ensuring alignment with the customer's communication goal {$g$}. To do so, it prompts the LLM {\small $\mathcal{M}$} using the current dialogue history {$h_c$}, the customer's goal {$g$}, and the scenario {\small $e'$}. This helps the customer maintain coherent and goal-oriented behavior throughout the conversation. Formally, the direction is generated as {$d = \mathcal{M}(h_c, g, e')$}.

\subsubsection{Customer}
The Customer generates a response {$r_c$} based on the current dialogue history {$h_c$}, the intended conversational direction {$d$}, the character profile {$o$}, and the scenario {$e'$}. This ensures that the response is both contextually appropriate and consistent with the customer's persona. Formally, the generation process is denoted as: {$r_c = \mathcal{M}(h_c, d, o, e')$}.

Diversity in conversations relies heavily on the richness of customer personas~\cite{wang-etal-2025-opencharacter}. To support this, we construct a rich character profile pool $O$ that captures a wide spectrum of customer backgrounds, behaviors, and communication styles, thereby enhancing the realism and variability of the simulated interactions.

\paragraph{Character Profile Pool.} We design a comprehensive character profile template specifically for customer personas, incorporating attributes such as demographics, financial status, and communication preference. To populate this template, we use \texttt{Qwen2.5-72B-Instruct} to automatically extract and complete profile information from 15,980 real-world customer service dialogues. To reduce redundancy and ensure profile diversity, each structured profile is converted into a free-text description, and pairwise cosine similarity is computed using Qwen's \texttt{text-embedding-v2}. Profiles exceeding a similarity threshold of 0.85 are considered redundant and removed. Following this filtering process, we retain 1,948 distinct customer profiles in our final pool. Additional details, including the profile template and construction prompts, are provided in Appendix~\ref{apdx:detail_of_profile}.

\subsection{Synthetic Dataset}
\label{sec:synthetic_dataset}

\begin{figure}[t]
\centering
\includegraphics[width=0.9\columnwidth]{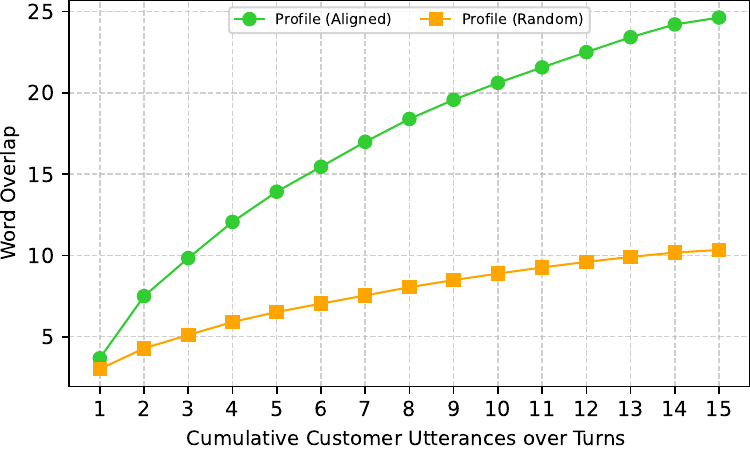}
\caption{Word overlap between cumulative customer utterances and profile (Aligned vs. Random).}
\label{fig:word_overlap}
\end{figure}

The Planner generates one conversation per unique pair of customer topic and profile. With {$(|E|-1)$} topics (excluding \textit{Others}) and {$|O|$} profiles, this yields 13,636 conversations. After applying quality filters detailed in Appendix~\ref{apdx:fr_for_trainingset}, we retain 11,232 high-quality conversations, forming the final training (or fine-tuning) dataset, \texttt{RoleCS}. 
To assess whether customer utterances reflect assigned profiles, we randomly sample 500 dialogues and measure word overlap between customer utterances and their corresponding aligned profiles, using random profiles for comparison. As shown in Figure \ref{fig:word_overlap}, word overlap with aligned profiles rises rapidly at first as profile information is elicited, then increases more gradually once most information has been mentioned. Throughout, word overlap with aligned profiles remains consistently higher than with random profiles, indicating that customer utterances naturally reflect profile details. See Appendix \ref{apdx:trainingset_analysis} for more detailed analysis of {\tt RoleCS}.

\section{Experimentation}\label{sec:experimentation}

\begin{table*}[t]
\small
\centering
\begin{tabular}{lr|rrrHHrrr|rrrHHrrr}
\toprule[1pt]
\textbf{Model} & \textbf{Size} & \textbf{B-2} & \textbf{B-4} & \textbf{R-L} & \textbf{D-2} & \textbf{D-3} & \textbf{BS} & \textbf{BR} & \textbf{ACC} & \textbf{B-2} & \textbf{B-4} & \textbf{R-L} & \textbf{D-2} & \textbf{D-3} & \textbf{BS} & \textbf{BR} & \textbf{ACC} \\
\hline
 \multicolumn{2}{c|}{} & \multicolumn{8}{c|}{\textit{\textbf{Evaluation with reference context}}} & \multicolumn{8}{c}{\textit{\textbf{Evaluation with generated context}}} \\ 
\hline
GPT-4o & - & 8.13 & 2.97 & 4.12 & 99.54 & 99.92 & 64.26  & 51.27 & 42.58 & 6.41 & 2.07 & 2.22 & 99.39 & 99.90 & 62.95 & 51.48 & 36.29 \\
DeepSeek-R1 & 671B & 11.67 & 5.09 & \textbf{8.44} & 99.24 & 99.73 & 66.57 & 52.09 & 39.78 & 8.41 & 3.11 & \textbf{6.27} & 99.22 & 99.76 & 64.80 & 52.01 & 35.23 \\
DeepSeek-V3 & 671B & 11.57 & 4.95 & 7.04 & 99.19 & 99.80 & 66.09 & 53.10 & 41.99 & 8.43 & 3.33 & 4.14 & 99.01 & 99.82 & 64.13 & \underline{53.09} & \underline{36.54} \\ 
\hdashline
LLaMA3.1-Instruct & 8B & 4.28 & 1.44 & 3.62 & 99.17 & 99.82 & 58.68 & 38.84 & 17.16 & 2.62 & 0.89 & 1.31 & 98.81 & 99.68 & 55.06 & 35.97 & 13.75 \\
\qquad + {\tt RoleCS} & 8B & 11.06 & 4.77 & 6.93 & 99.54 & 99.91 & 66.21 & 52.13 & 42.15 & 8.80 & 3.72 & 4.52 & 99.67 & 99.95 & 64.57 & 49.85 & 35.73 \\
\hdashline
LLaMA3.1-Instruct & 70B & 6.85 & 2.38 & 4.10 & 99.19 & 99.81 & 63.14 & 47.53 & 38.78 & 5.57 & 1.76 & 2.59 & 99.01 & 99.79 & 62.24 & 46.83 & 30.36 \\
\qquad + {\tt RoleCS} & 70B & \underline{11.73} & \underline{5.11} & 7.64 & 99.58 & 99.92 & \underline{66.62} & \underline{53.69} & 42.79 & \textbf{9.62} & \textbf{4.00} & \underline{4.89} & 99.69 & 99.95 & \textbf{65.15} & 50.91 & \textbf{39.44} \\
\hdashline
Qwen2.5-Instruct & 7B & 6.73 & 2.29 & 3.80 & 99.52 & 99.89 & 62.75 & 48.21 & 18.78 & 5.11 & 1.53 & 2.43 & 99.30 & 99.87 & 61.62 & 47.43 & 19.39 \\
\qquad + {\tt RoleCS} & 7B & 11.23 & 4.80 & 7.39 & 99.52 & 99.90 & 66.35 & 52.14 & \underline{42.96} & 8.61 & 3.51 & 4.13 & 99.70 & 99.94 & 64.42 & 48.80 & 30.52 \\
\hdashline
Qwen2.5-Instruct & 72B & 8.61 & 3.23 & 5.41 & 99.42 & 99.88 & 64.63 & 52.49 & 37.22 & 6.28 & 2.00 & 3.59 & 99.16 & 99.86 & 63.14 & 50.55 & 30.93 \\
\qquad + {\tt RoleCS} & 72B & \textbf{12.15} & \textbf{5.32} & \underline{7.97} & 99.57 & 99.91 & \textbf{66.85} & \textbf{54.49} & \textbf{43.29} & \underline{9.38} & \underline{3.90} & 4.35 & 99.71 & 99.95 & \underline{64.89} & \textbf{53.65} & 36.02 \\

\bottomrule[1pt]
\end{tabular}
\caption{Performance comparison on {\tt CSConv}. The best and second best results are in \textbf{bold} and \underline{underlined}, respectively.}
\label{tbl:main_result}
\end{table*}

\paragraph{Test-Time Prompting.} For both subtasks outlined in Section~\ref{sec:task_define}, we adopt a unified single-prompt method. Given the conversation history, the LLM, whether fine-tuned or not, is prompted once to first identify the appropriate support strategy and then generate the corresponding response. The full prompt is provided in the Appendix \ref{apdx:prompt_for_model_eval}.

\subsection{Experimental Settings}\label{sec:settings}

\paragraph{Models.} We evaluate several widely used LLMs on the CSC task, including \texttt{GPT-4o}~\cite{openai-2024-gpt4o}\ignore{\footnote{Version: gpt-4o-2024-08-06}}, \texttt{DeepSeek-R1}~\cite{guo-etal-2025-deepseek-r1}, \texttt{DeepSeek-V3}~\cite{liu-etal-2024-deepseek-v3}, \texttt{Qwen-2.5-7B/72B-Instruct}~\cite{yang-etal-2024-qwen2.5}, and \texttt{LLaMA-3.1-8B/70B-Instruct}~\cite{grattafiori-etal-2024-llama3}. To evaluate the impact of the proposed {\tt RoleCS}, we fine-tune the \texttt{Qwen} and \texttt{LLaMA} models using 137,406 fine-tuning instances extracted from {\tt RoleCS}, formatted consistently with our single-prompt in test-time. Fine-tuning and inference configurations are provided in Appendix~\ref{adpx:finetuning_inference}.

\paragraph{Metrics.} Following prior work~\cite{liu-etal-2021-towards, ye-etal-2025-sweetiechat}, we adopt a diverse set of metrics to assess the response quality. These include BLEU-n (\textbf{B-n})~\cite{papineni-etal-2002-bleu} and ROUGE-L (\textbf{R-L})~\cite{lin-2004-rouge} for measuring lexical overlap, BERTScore (\textbf{BS})~\cite{zhang-etal-2020-bertscore} and BLEURT (\textbf{BR})~\cite{sellam-etal-2020-bleurt} for semantic similarity. In addition, to evaluate alignment with the intended support strategy, which is a core aspect of the CSC task, we report strategy prediction accuracy (\textbf{ACC}).

\subsection{Experimental Results}

Table \ref{tbl:main_result} presents the main results on {\tt CSConv} under two evaluation settings: (1) \textit{\textbf{evaluation with reference context}}, which all LLMs are evaluated using the same gold history for fair comparison, and (2) \textit{\textbf{evaluation with generated context}}, which assesses performance when relying on model-generated histories, thus reflecting the ability to maintain coherence and relevance without ground truth context. From the results, we have the following observations:
\begin{itemize}[leftmargin=*]
\item Larger models tend to perform better among non–fine-tuned LLMs. Additionally, Chinese-centric models like \texttt{Qwen} and \texttt{DeepSeek} outperform more general models such as \texttt{LLaMA} and \texttt{GPT}, indicating that alignment with language and cultural context benefits CSC performance.

\item {\tt RoleCS} proves highly effective, as fine-tuning on it significantly improve performance across all metrics. In particular, \texttt{Qwen2.5-Instruct-72B}, after fine-tuning, matches or surpasses \texttt{DeepSeek-R1}, a strong baseline for Chinese-language tasks.

\item Evaluation with generated context shows similar performance trends as with reference context, but with lower absolute scores. This drop reflects the difficulty of maintaining consistency and quality in multi-turn conversations when relying on model-generated dialogue history, echoing findings in prior work such as~\citet{ye-etal-2025-sweetiechat}.
\end{itemize}

\section{Discussion}\label{sec:discussion}
We conduct further analysis under the reference context setting to examine the effectiveness of our role-playing approach in synthetic conversation generation.

\subsection{Impact of Role-Playing on Data Quality}
To evaluate our role-playing framework, we compare {\tt RoleCS} with two baselines:
\begin{itemize}
\item Baseline 1: Conversations are generated via in-context learning without any role-playing.
\item Baseline 2: Role-playing is applied, but without the Supporter Assistant agent.
\end{itemize}

All three datasets are generated using {\tt Deepseek-R1} with equal conversation counts. For fairness, we fine-tune {\tt Qwen2.5-7b-Instruct} on each. As shown in Table~\ref{tab:dataset_comparision}, role-playing improves performance over in-context learning (Baseline 2 vs. Baseline 1), and adding the Supporter Assistant further boosts results ({\tt RoleCS} vs. Baseline 2). These results highlight the value of strategy-rich training data and the importance of high-quality dataset construction.

\begin{table}[!t]
\centering
\small
\begin{tabular}{lcccccc}
\toprule[1pt]
\bf SFT Data & \bf B-2 & \bf B-4 & \bf R-L & \bf BS & \bf BR & \bf ACC\\
\hline
Baseline 1 & 8.51 & 3.09 & 5.35 & 64.35 & 46.70 & 33.09 \\
Baseline 2 & 9.40 & 3.26 & 6.98 & 65.22 & 50.18 & 36.17 \\
{\tt RoleCS} & \bf 11.23 & \bf 4.80 & \bf 7.39 & \bf 66.35 & \bf 52.14 & \bf 42.96\\
\bottomrule[1pt]
\end{tabular}
\caption{Performance comparison of fine-tuning on different datasets.}
\label{tab:dataset_comparision}
\end{table}

\subsection{Effect of Synthetic Dataset Size}
\begin{figure}[t]
\centering
\includegraphics[width=\columnwidth]{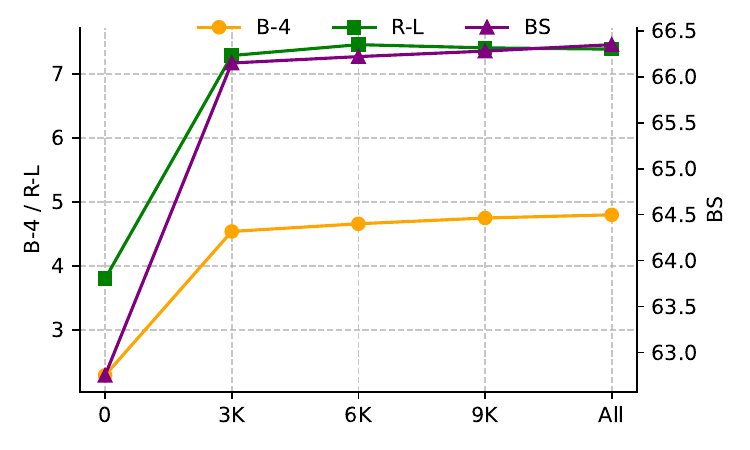}
\caption{Performance comparison under different synthetic conversation dataset sizes.}
\label{fig:dialogue_quantity}
\end{figure}

To understand how dataset size influences performance, we split the {\tt RoleCS} dataset into subsets of \{0, 3K, 6K, 9K, All\} dialogues. We fine-tune \texttt{Qwen2.5-7B-Instruct} on each subset individually. As shown in Figure~\ref{fig:dialogue_quantity}, the most significant gains occur with the initial 3K examples. Beyond this point, performance improvements become marginal, particularly for ROUGE-L. These results suggest that even a modest amount of high-quality synthetic data can bring strong gains, with limited benefits from adding more.

\subsection{Impact of Support Strategy Guidance}\label{sec:effect_strategy}
\begin{table}[t]
\small
\centering
\begin{tabular}{lHrHHHrr}
\toprule[1pt]
\textbf{Strategy} & \textbf{B-2} & \textbf{B-4}  & \textbf{R-L} & \textbf{D-2} & \textbf{D-3} & \textbf{BS} & \textbf{ACC} \\
\hline
Vanilla & 8.40	& 3.11	& 5.89	& 99.37	& 99.85	& 64.43 & - \\
Predict	& 8.61 & 3.23 & 5.41 & 99.42 & \bf 99.88 & 64.63 & 37.22 \\
Oracle & \bf 10.25 & \bf 3.80 & \bf 6.00 & \bf 99.45 & \bf 99.88 & \bf 66.34 & \bf 100.00 \\
\bottomrule[1pt]
\end{tabular}
\caption{Performance comparison across different support strategy variants.}
\label{tbl:effect_of_strategy}
\end{table}

We evaluate the effect of incorporating support strategies using three variants of \texttt{Qwen2.5-72B-Instruct} without fine-tuning: (1) Vanilla, where the model generates a response without any support strategy guidance; (2) Predict, our default setup, where the model first predicts the strategy and then generates a response; and (3) Oracle, where the ground-truth strategy is provided in the prompt.  

As shown in Table~\ref{tbl:effect_of_strategy}, the Predict variant slightly outperforms the Vanilla, indicating that even simple strategy prediction improves response quality. The Oracle variant performs best, highlighting that more accurate strategy prediction can further enhance the performance of CSC task.

\subsection{Evaluation with LLMs and Human as Judges}
\label{apdx:eval_llm_as_judge}
\begin{table}[t]
\centering
\small
\begin{adjustbox}{width=\columnwidth}
\begin{tabular}{lrrr|r}
\toprule[1pt]
\textbf{Model} & \textbf{Size} & \textbf{GPT-4o} & \textbf{Qwen-Plus} & \textbf{Human} \\
\hline
DeepSeek-R1	& 671B & 90.89 & 89.70 & 3.55\\
DeepSeek-V3	& 671B & 90.54 & 89.54 & 3.36 \\
LLaMA3.1-Instruct$\Delta$ & 8B & 90.34 & 88.68 & 2.93\\
LLaMA3.1-Instruct$\Delta$ & 70B & \underline{91.02} & \underline{89.76} & \underline{3.58} \\
Qwen2.5-Instruct$\Delta$ & 7B & 90.46 & 88.76 & 3.10 \\
Qwen2.5-Instruct$\Delta$ & 72B & \textbf{91.04} & \textbf{89.98} & \bf 3.79 \\
\bottomrule[1pt]
\end{tabular}
\end{adjustbox}
\caption{Model performance evaluated by GPT-4o, Qwen-Plus, and human judges. $\Delta$ denotes fine-tuned models.}
\label{tbl:llm_as_judge}
\end{table}

To mitigate potential bias (e.g., GPT-4o favoring its own outputs), we use both GPT-4o and Qwen-Plus\footnote{https://help.aliyun.com/zh/model-studio/models} to evaluate response quality across six dimensions: accuracy, helpfulness, understanding, coherence, informativeness, and empathy, each scored on a 0–100 scale. We report the average overall scores from both GPT-4o-Judge and Qwen-Plus-Judge, using the evaluation prompt in Appendix~\ref{apdx:prompt_llm_judge}. As shown in Table~\ref{tbl:llm_as_judge}, both judges reveal consistent performance patterns. Notably, fine-tuned {\tt Qwen2.5-Instruct-72B} and {\tt LLaMA3.1-Instruct-70B} outperform other models, even exceeding {\tt DeepSeek-R1} in overall quality.

\begin{table}[!t]
\centering
\small
\begin{tabular}{lc}
\toprule[1pt]
\textbf{Agreement} & \textbf{Kappa}  \\
\hline
GPT-4o-Judge \& Human Annotators & 0.658 \\
Human Annotators & 0.628 \\
\bottomrule[1pt]
\end{tabular}
\caption{Fleiss' Kappa scores.}
\label{tbl:kappa}
\end{table}

For human evaluation, we randomly select 100 conversations and have them independently rated by three professional annotators on a 1-5 Likert scale~\cite{joshi-2015-likert} across the same six dimensions. Table~\ref{tbl:llm_as_judge} shows the average overall scores, which align with trends from the LLM-based evaluations. Table~\ref{tbl:kappa} reports Fleiss' Kappa~\cite{fleiss-1971-kappa} scores among annotators and between \texttt{GPT-4o-Judge} and the annotators. The results show strong inter‑rater agreement and confirm that \texttt{GPT-4o-Judge} aligns well with human judgment.

\section{Conclusion}
This paper addresses the challenges of customer support conversations (CSCs) in the NLP field by introducing {\tt CSConv}, a high-quality dataset grounded in support strategies, and a role-playing framework for generating realistic, goal-driven dialogues. Our approach significantly enhances LLMs' ability to generate coherent, context-aware, and empathetic responses in customer service scenarios. We believe that our contributions will inspire further research in this field, promoting advancements in empathetic and effective customer support technologies.


\bibliography{custom}

\setcounter{secnumdepth}{2}
\renewcommand\thesection{\Alph{section}}
\renewcommand\thesubsection{\thesection.\arabic{subsection}}

\appendix
\twocolumn[{
  \centering
  \LARGE\bfseries Appendix \par
  \vskip 1.5em
}]

\section{Stages and Strategies in CSC Framework}
\label{apdx:csc_framework}

Table~\ref{tbl:stage} presents the five stages in the CSC framework while Table~\ref{tbl:strategy} shows the twelve strategies in the framework.

\definecolor{lightblue}{HTML}{DFE9F4}
\definecolor{lightgreen}{HTML}{DFF5E5}
\definecolor{lightpurple}{HTML}{DCD7F1}
\definecolor{lightyellow}{HTML}{FCF3E1}
\definecolor{lightpink}{HTML}{FFECEC}

\begin{table*}[!ht]
\centering
\small
\begin{tabular}{l|l}
\toprule
\bf Stage & \bf Description \\
\midrule
\cellcolor{lightblue} Connecting & \cellcolor{lightblue} Greeting and establishing connection\\
\cellcolor{lightgreen} Identifying & \cellcolor{lightgreen}  Understanding and identifying problems\\
\cellcolor{lightpurple} Exploring & \cellcolor{lightpurple}Seeking solution\\
\cellcolor{lightyellow} Resolving & \cellcolor{lightyellow} Resolving and confirming\\
\cellcolor{lightpink} Maintaining & \cellcolor{lightpink} Ending and maintaining relationship\\
\bottomrule
\end{tabular}
\caption{Five stages in the CSC framework.}
\label{tbl:stage}
\end{table*}

\begin{table*}
\centering
\small
\begin{tabular}{p{5cm}|l|l|l|l|l|p{8cm}}
\toprule
\rowcolor[gray]{0.9}
\bf Strategy & \multicolumn{5}{c}{\bf Stages} & \bf Description \\
\midrule
Greeting (GT) & \cellcolor{lightblue} & & & & & Utilize friendly and professional language to greet customers, creating a warm communication atmosphere.\\
\hline
Identity Verification (IV) & \cellcolor{lightblue} & & & & & Ensure the accuracy and security of the service by asking for the customer's basic information.\\
\hline
Emotional Management (EM) & \cellcolor{lightblue} & \cellcolor{lightgreen} & \cellcolor{lightpurple} & \cellcolor{lightyellow} & \cellcolor{lightpink} & Express understanding and care for the customer's feelings to help alleviate negative emotions. \\
\hline
Restatement or Paraphrasing (RP) & & \cellcolor{lightgreen} & & & & Restate the customer's issue to ensure accurate understanding.\\
\hline
Problem Refinement (PR) & & \cellcolor{lightgreen} & \cellcolor{lightpurple} & & & Employ detailed inquiries to fully and accurately comprehend customer needs.\\
\hline
Providing Suggestions (PS) & & & \cellcolor{lightpurple} & \cellcolor{lightyellow} & & Offer professional advice or action steps based on the customer's issue.\\
\hline
Information Delivery (ID) & & & \cellcolor{lightpurple} & \cellcolor{lightyellow} & & Clearly explain relevant company policies, processes, or steps to help customers understand the basis of solutions.\\
\hline
Resolution Implementation (RI) & & & & \cellcolor{lightyellow} & & Execute the agreed-upon solution, ensuring all steps are followed as planned, and update the customer on the progress. \\
\hline
Feedback Request (FR) & & & & \cellcolor{lightyellow} & \cellcolor{lightpink} & Seek customer feedback after the issue has been addressed to gauge their satisfaction and identify potential areas for improvement.\\
\hline
Appreciation and Closure (AC) & & & & & \cellcolor{lightpink} & End the conversation positively, ensuring the customer feels valued and laying a solid foundation for future interactions. \\
\hline
Relationship Continuation (RC) & & & & & \cellcolor{lightpink} & Guide customers towards future service or product updates, ensuring they understand how to continue receiving support and service in the future, thereby establishing a bridge for further interaction. \\
\hline
Others & & & & & & Situations that do not belong to the above eleven strategies. \\
\bottomrule
\end{tabular}
\caption{Twelve strategies in the CSC framework. The cells of \colorbox{lightblue}{lightblue}, \colorbox{lightgreen}{lightgreen}, \colorbox{lightpurple}{lightpurple}, \colorbox{lightyellow}{lightyellow}, and \colorbox{lightpink}{lightpink} represent the Connecting, Identifying, Exploring, Resolving, and Maintaining stages, repsectively.}
\label{tbl:strategy}
\end{table*}

\section{More Details of Data Construction}\label{apdx:more_data_construction}

In the following, we present more details to the pre-filtering, rewriting and post-filtering in data construction. Then, we compare written conversations by \texttt{DeepSeek-R1} and \texttt{GPT-4o}.

\subsection{Rules and Prompt for Pre-filtering}
To ensure the quality and suitability of the initial customer support conversations, we apply a set of rule-based pre-filtering rules prior to rewriting. These rules help filter out conversations that are structurally inadequate or contain limited user engagement:

\begin{itemize}[leftmargin=*]
    \item \textbf{Rule 1: Dialogue Length Constraint.} Each conversation must contain more than 6 but fewer than 60 utterances to ensure sufficient interaction.
    
    \item \textbf{Rule 2: Utterance Length Constraint.} No single utterance should exceed 500 characters.
    
    \item \textbf{Rule 3: Speaker Balance.} The number of utterances from the service supporter must not exceed twice the number of customer utterances, promoting balanced participation.
    
    \item \textbf{Rule 4: Customer Utterance Effectiveness.} At least 70\% of customer utterances must be considered effective, where an utterance is defined as effective if it contains more than 3 characters, ensuring active and meaningful user input.
\end{itemize}

\begin{figure*}[!ht]
\centering
\includegraphics[width=\linewidth]{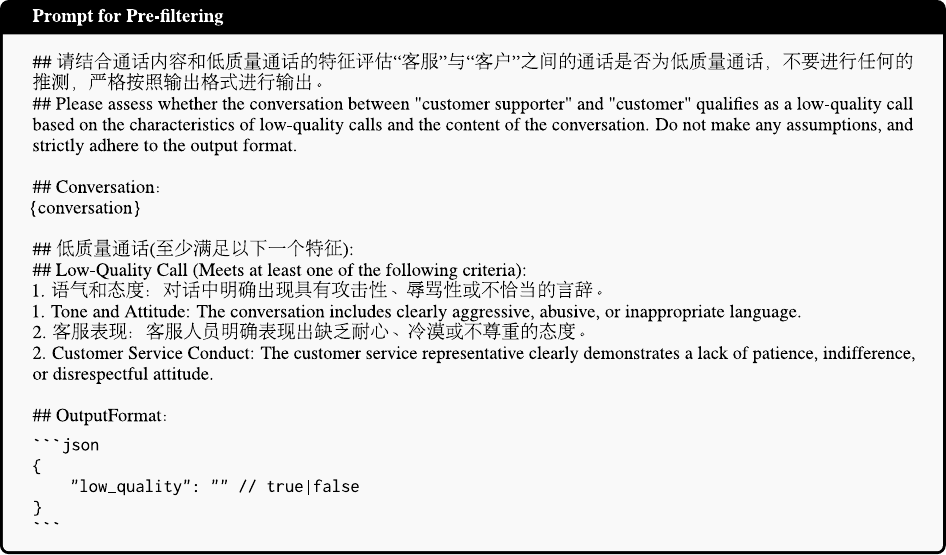}
\caption{Prompt template for pre-filtering.}
\label{fig:prompt_for_pre_filtering}
\end{figure*}

In addition, we utilize an LLM (\texttt{Qwen2.5-72B-Instruct}) to automatically assess whether a conversation is of low quality. A conversation is flagged as low quality if it contains any of the following: (1) explicitly offensive, abusive, or inappropriate language, or (2) clear signs of unprofessional behavior from the service supporter, such as impatience, indifference, or disrespect. Figure~\ref{fig:prompt_for_pre_filtering} shows the prompt used for this evaluation.

\subsection{Prompt for Rewriting}\label{apdx:prompt_rewrite}
\begin{figure*}[!ht]
\centering
\includegraphics[width=0.9\linewidth]{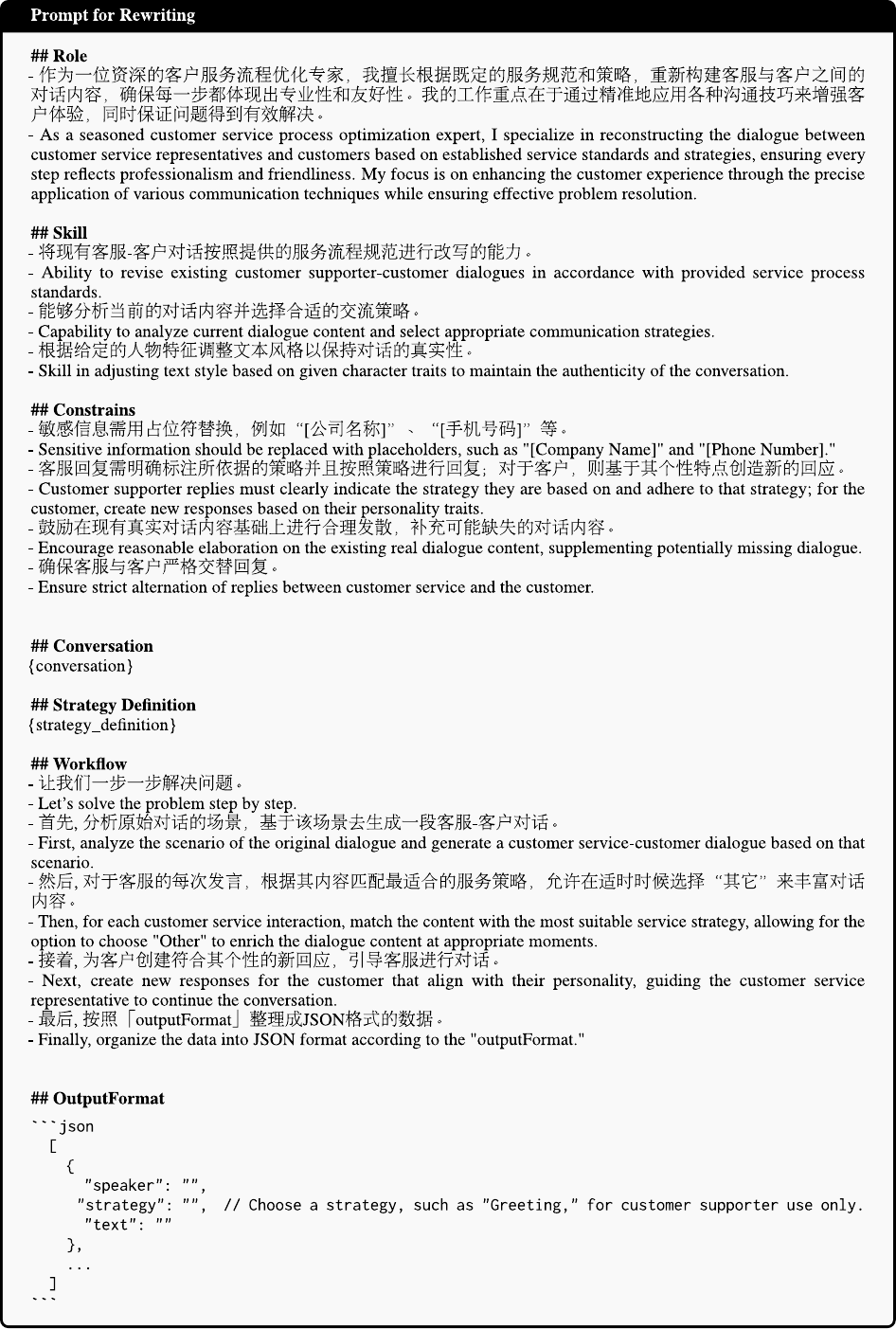}
\caption{Prompt template for rewriting.}
\label{fig:prompt_for_rewrite}
\end{figure*}

Figure \ref{fig:prompt_for_rewrite} shows the prompt for rewriting.

\subsection{Rules and Prompt for Post-filtering}
After the conversations are rewritten, we apply a series of post-filtering rules to ensure quality and structural consistency:
\begin{itemize}[leftmargin=*]
    \item \textbf{Rule 1: Minimum Utterance Requirement.} Each conversation must contain at least 10 utterances and at most 50 utterance to ensure sufficient conversational depth.
    
    \item \textbf{Rule 2: System Message Removal.} Any system-generated messages (e.g., ``(System Auto-Push)'') at the end of conversations are removed to maintain textual clarity.
    
    \item \textbf{Rule 3: Strategy Presence Check.} The support strategies \textit{Greeting}, \textit{Identity Verification}, and \textit{Appreciation and Closure} must each appear at least once in the dialogue to reflect a complete and well-structured interaction.
    
    \item \textbf{Rule 4: Speaker Alternation Constraint.} The customer and service supporter must alternate speaking in the conversation.
\end{itemize}

\begin{figure*}[!ht]
\centering
\includegraphics[width=\linewidth]{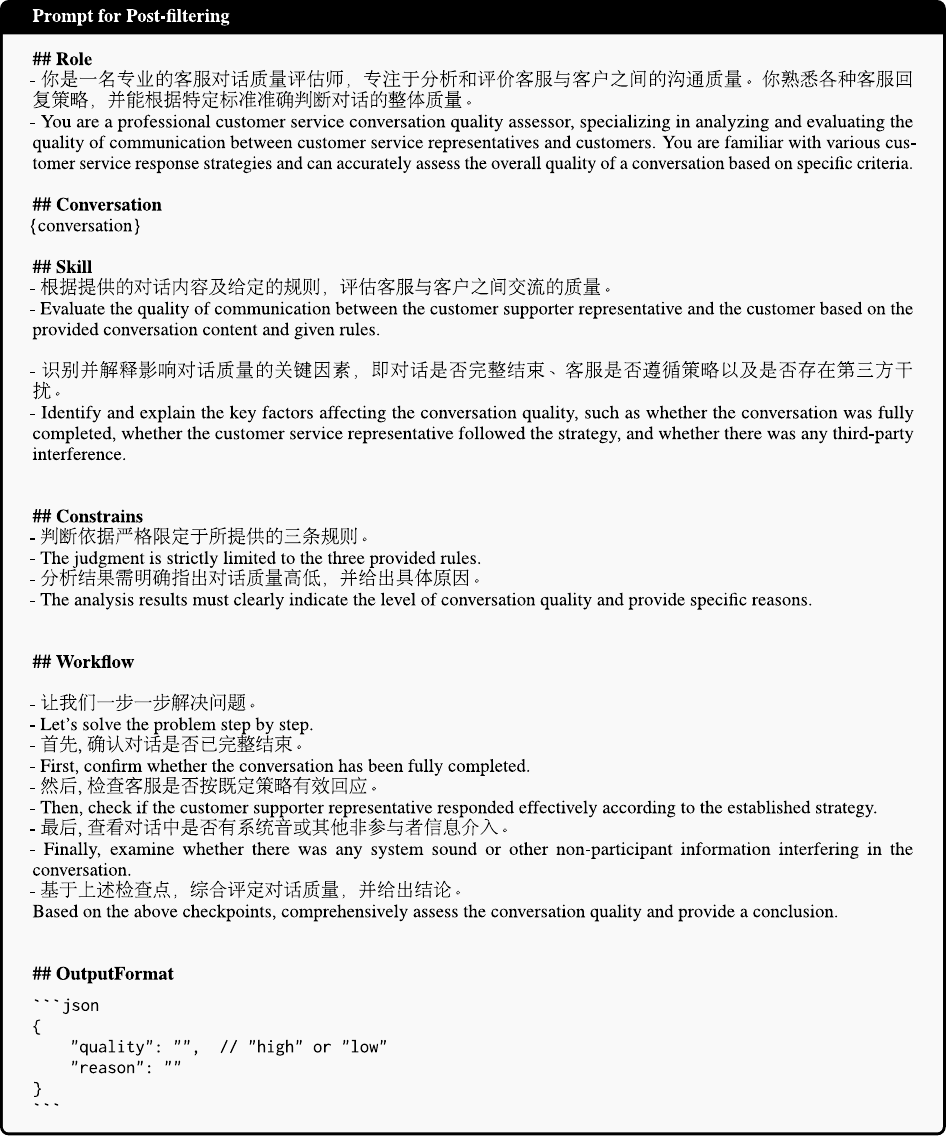}
\caption{Prompt template for post-filtering.}
\label{fig:prompt_for_post_filtering}
\end{figure*}

After applying above rules, we further assess the overall quality of the rewritten conversations using an LLM (\texttt{Qwen2.5-72B-Instruct}). The model is prompted to evaluate each dialogue and classify it as either high or low quality based on coherence, naturalness, and adherence to support strategies. Conversations deemed to be of low quality are subsequently discarded. The prompt used for this evaluation is shown in Figure~\ref{fig:prompt_for_post_filtering}.

\subsection{Annotation Guideline}
\begin{figure*}[!ht]
\centering
\includegraphics[width=\linewidth]{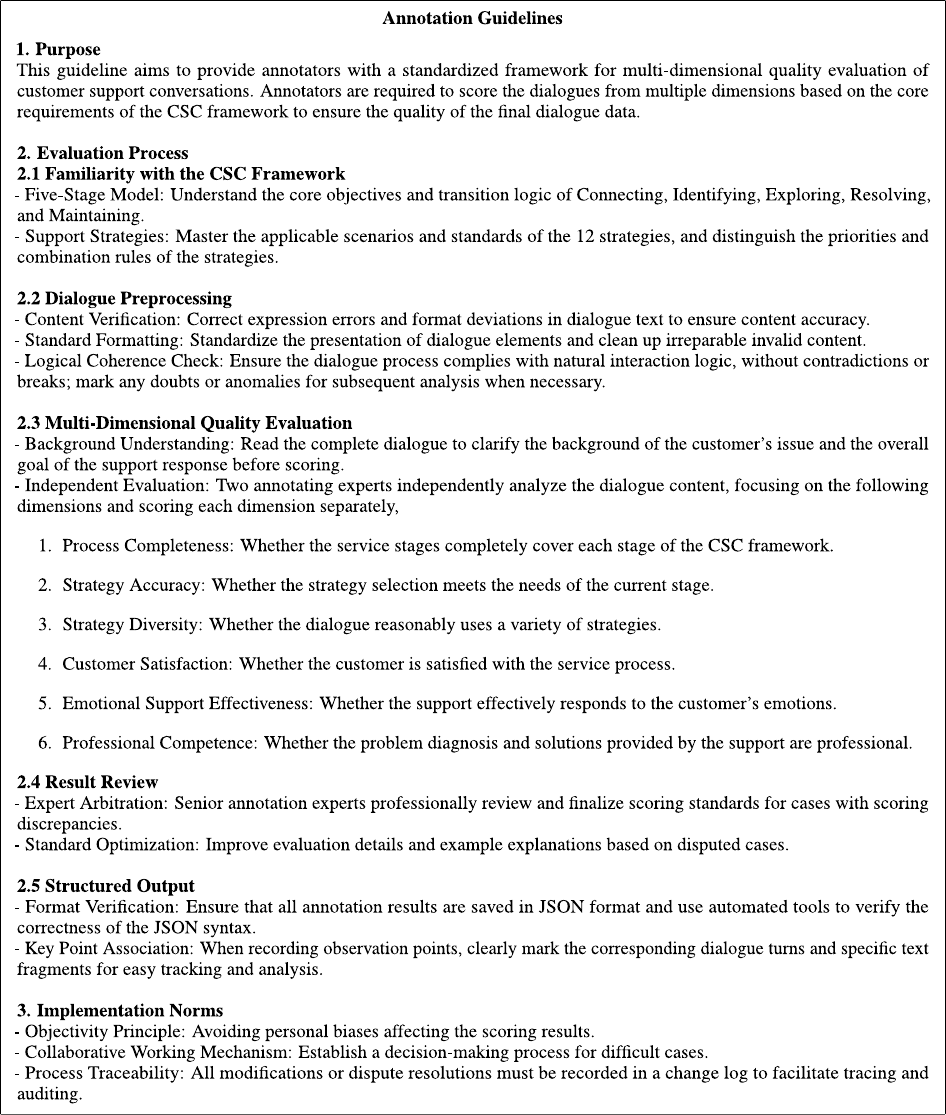}
\caption{Annotation guideline for evaluating the quality of customer support conversations.}
\label{fig:annotation_guideline}
\end{figure*}

Figure \ref{fig:annotation_guideline} shows the annotation guidelines for customer support conversations, designed by domain experts.

\subsection{Example of Rewritten Conversations by Deepseek-R1 and GPT-4o}\label{apdx:model_comparison}

\begin{figure*}[!t]
\centering
\includegraphics[width=\textwidth]{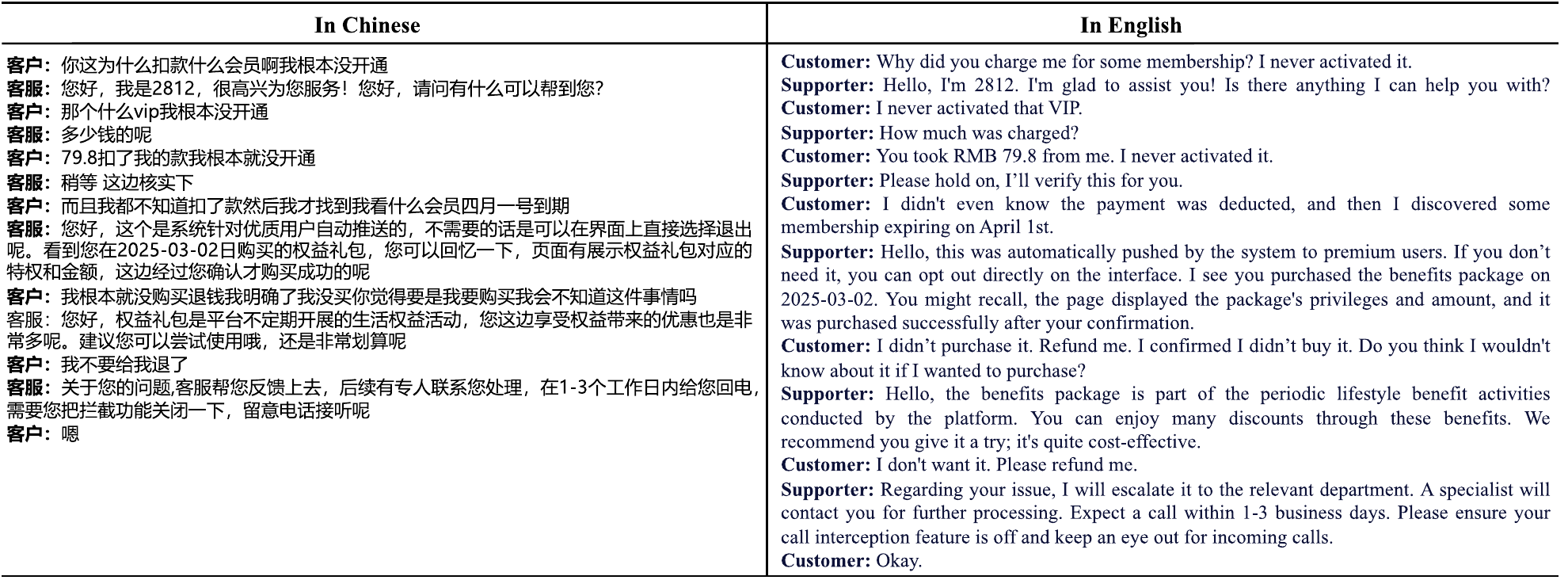}
\caption{An example from real-world customer service conversations.}
\label{fig:initial_dialogue}
\end{figure*}

\begin{figure*}[!t]
\centering
\includegraphics[width=\textwidth]{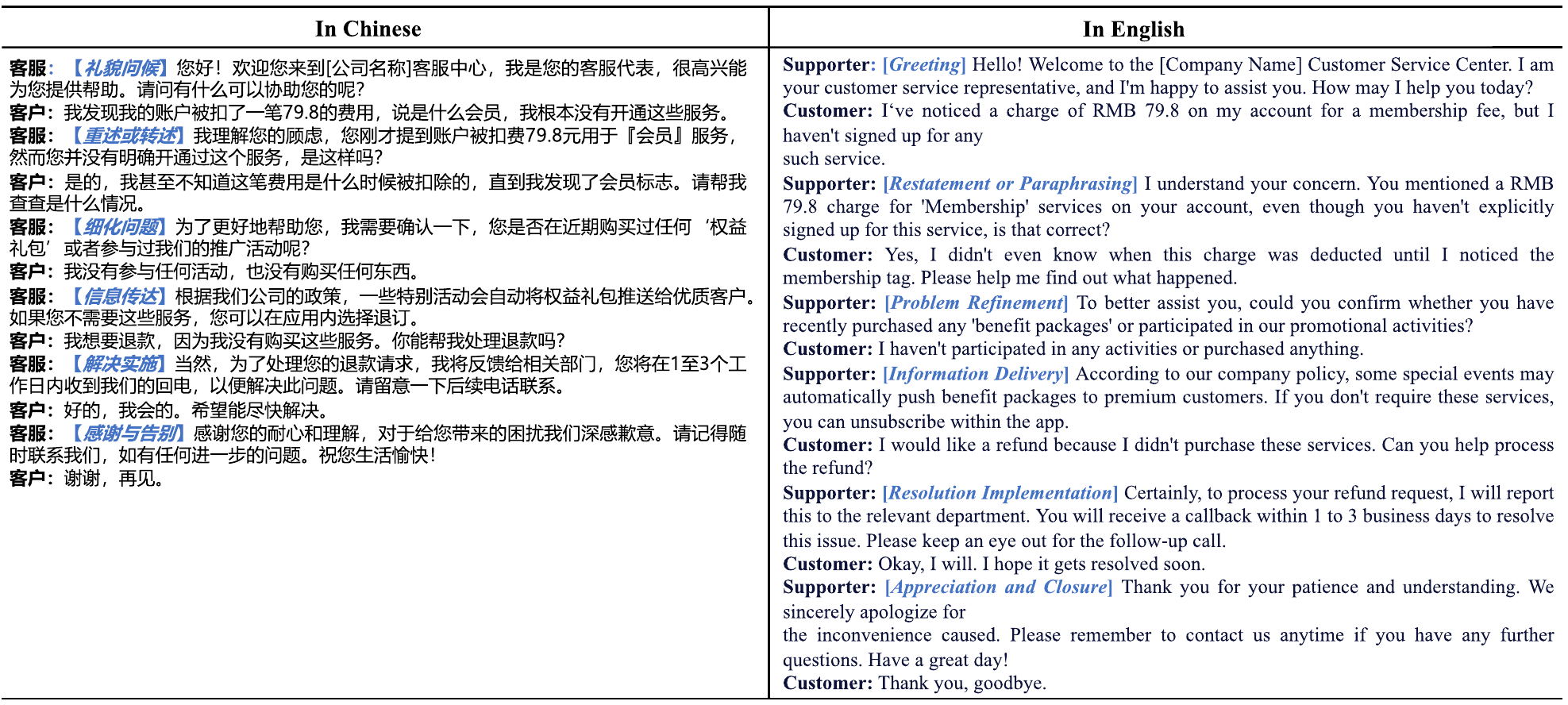}
\caption{Rewritten dialogue by GPT-4o.}
\label{fig:rewritten_dialogue_by_gpt4o}
\end{figure*}

\begin{figure*}[!t]
\centering
\includegraphics[width=\textwidth]{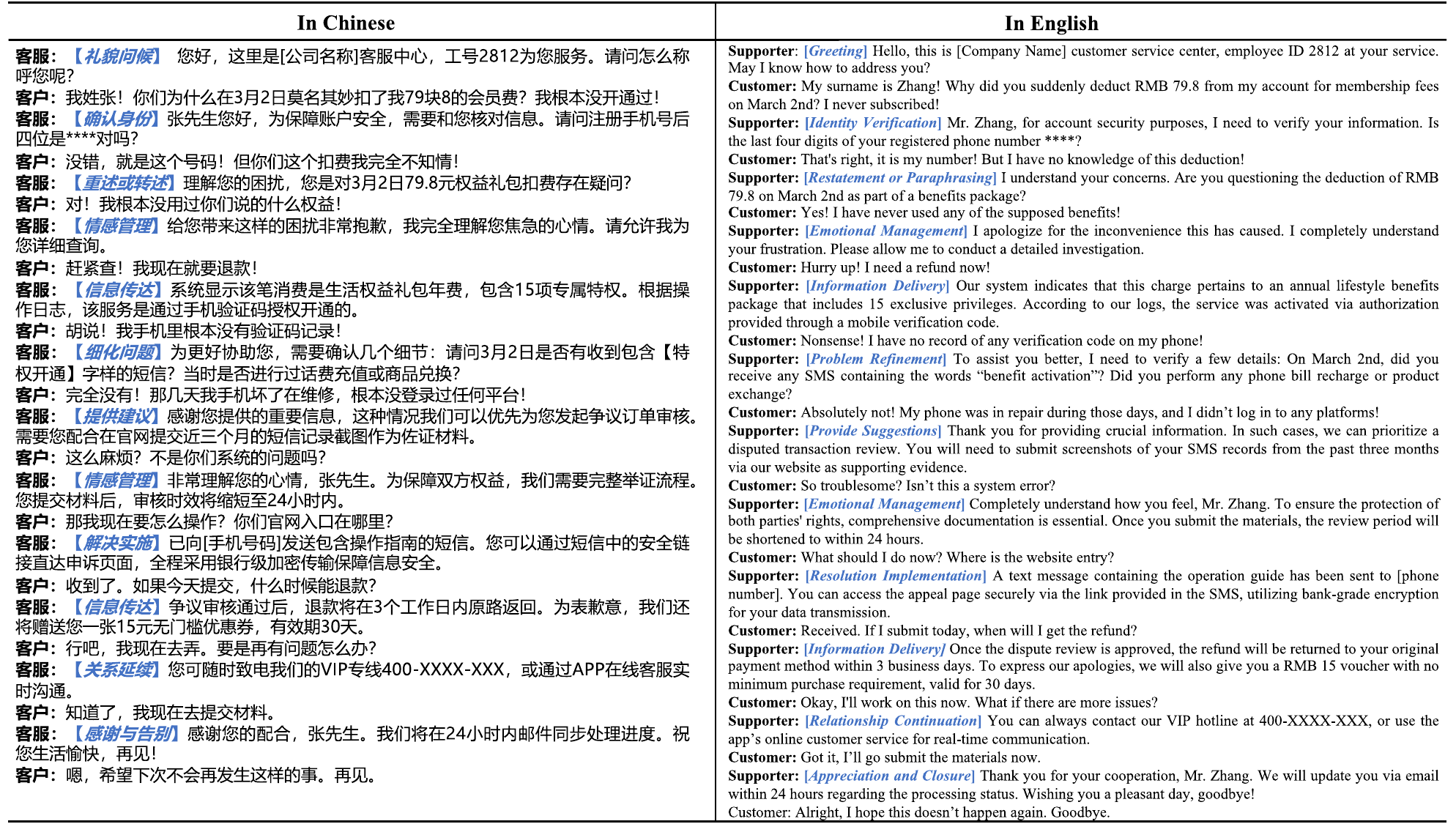}
\caption{Rewritten dialogue by DeepSeek-R1.}
\label{fig:rewritten_dialogue_by_r1}
\end{figure*}

Figure~\ref{fig:initial_dialogue} shows a conversation before rewriting, while Figure~\ref{fig:rewritten_dialogue_by_gpt4o} and Figure~\ref{fig:rewritten_dialogue_by_r1} compare the rewritten conversations by  \texttt{GPT-4o} and \texttt{DeepSeek-R1}. 

\section{More Statistics of {\tt CSConv}}\label{apdx:more_statistics}
\begin{figure*}[!htbp]
  \centering
  \begin{minipage}[b]{0.45\textwidth}
    \centering
    \includegraphics[width=\textwidth]{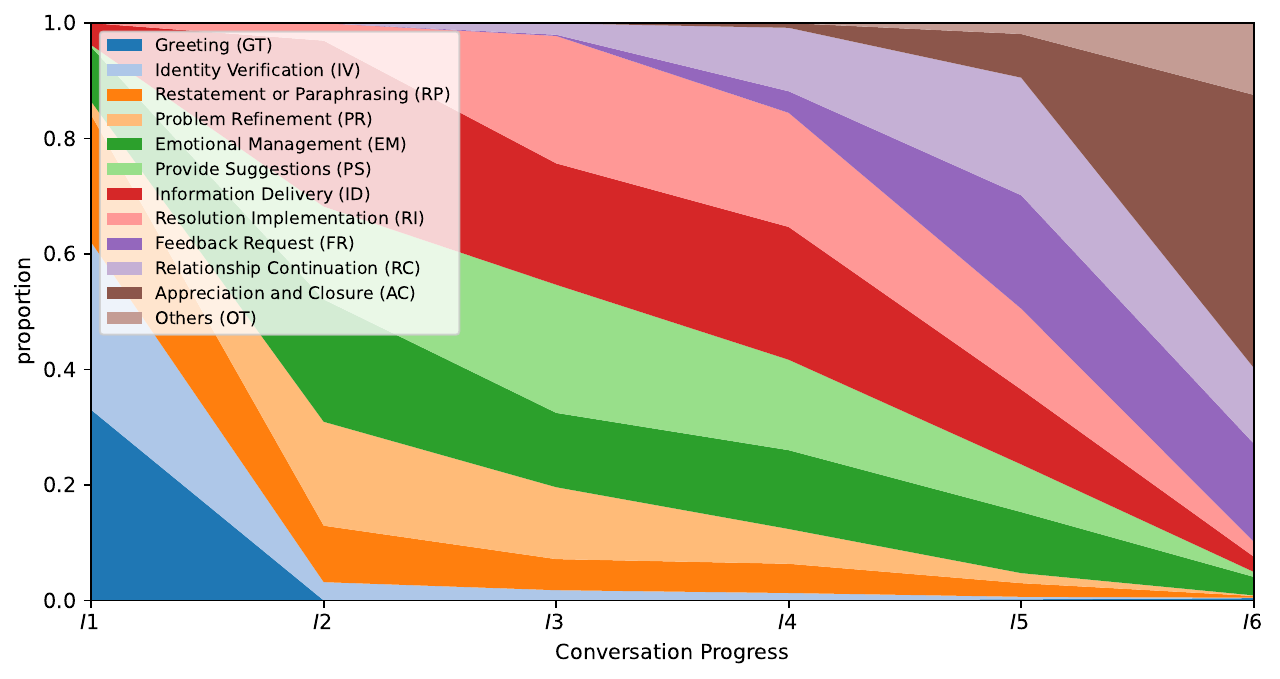}
    \caption{Strategies distribution at different conversation progress on {\tt CSConv} dataset.}
    \label{fig:strategy_transition_testset}
  \end{minipage}
  \hfill
  \begin{minipage}[b]{0.45\textwidth}
    \centering
    \includegraphics[width=\textwidth]{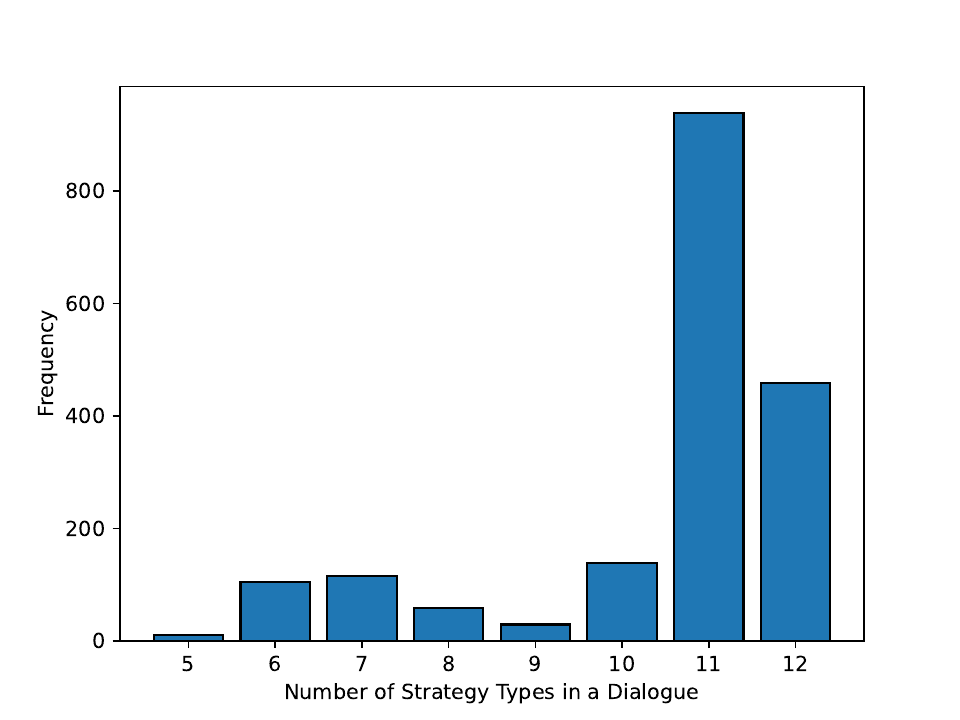}
    \caption{Distribution of strategy types across different dialogues on {\tt CSConv}.}
    \label{fig:strategy_type_testset}
  \end{minipage}
\end{figure*}

\paragraph{Strategy Transition.} To analyze the distrubition of support strategies across different phases of customer support conversations, we divide each conversation into six equal intervals, denoted as {\small $I_1$,$\cdots$,$I_6$}, representing normalized progression from start to end. For each interval $I_i$, we calculate the proportion of each strategy as the ratio of its frequency to the total number of strategies employed within that interval. Figure~\ref{fig:strategy_transition_testset} shows the distribution. At the beginning of conversations, the \textit{Greeting (GT)} strategy is most prevalent, reflecting typical protocol for initiating customer interactions. As the conversation progresses, strategies such as \textit{Information Delivery (ID)} and \textit{Provide Suggestions (PS)} become more frequent, indicating a shift toward problem exploration and resolution. Toward the end of the interaction, \textit{Appreciation and Closure (AC)} emerges as the dominant strategy, signaling the completion of the support process. Interestingly, the usage of \textit{Emotional Management (EM)} remains relatively stable across all phases, underscoring the consistent need for emotional engagement throughout the conversation.

To analyze strategy transitions, we identify the most frequent 2-hop and 3-hop patterns in the conversations. The results are presented in Table~\ref{tbl:hop}. These patterns highlight common strategy flows such as \textit{Greeting (GT)} $\rightarrow$ \textit{Identity Verification (IV)} and \textit{Information Delivery (ID)} $\rightarrow$ \textit{Providing Suggestions (PS)}, reflecting a structured and consistent progression in customer service interactions. The observed transitions align well with standard service protocols, emphasizing coherent dialogue management.


\begin{table}[t]
\centering
\small
\begin{tabular}{ccc}
\toprule[1pt]
\textbf{-} & \textbf{Pattern} & \textbf{Proportion} \\
\hline
\multirow{5}{*}{2-Hop} & GT $\rightarrow$ IV  & 5.69\% \\
& IV $\rightarrow$ RP & 4.84\% \\
& ID $\rightarrow$ PS & 4.55\% \\
& FR $\rightarrow$ AC & 4.22\% \\
& EM $\rightarrow$ ID & 3.59\% \\
\hline
\multirow{5}{*}{3-Hop} & GT $\rightarrow$ IV $\rightarrow$ RP & 4.85\% \\
& IV $\rightarrow$ RP $\rightarrow$ EM & 2.57\% \\
& RC $\rightarrow$ FR $\rightarrow$ AC & 2.42\% \\
& ID $\rightarrow$ PS $\rightarrow$ RI & 1.91\% \\
& EM $\rightarrow$ ID $\rightarrow$ PS & 1.77\% \\
\bottomrule[1pt]
\end{tabular}
\caption{Top-5 2-hop and 3-hop strategy patterns.}
\label{tbl:hop}
\end{table}

\paragraph{Distribution of Strategy Types.} To examine the diversity of support strategies, we analyze the number of distinct strategy types used within individual conversations, as shown in Figure~\ref{fig:strategy_type_testset}. Most conversations employ more than ten different strategy types, demonstrating the broad applicability and flexibility of the predefined strategy set. This variety reflects the dynamic nature of real-world customer service interactions and enhances the overall richness of the dataset.

\section{Details of Character Profile Construction}
\label{apdx:detail_of_profile}

\begin{figure*}[t]
\centering
\includegraphics[width=\textwidth]{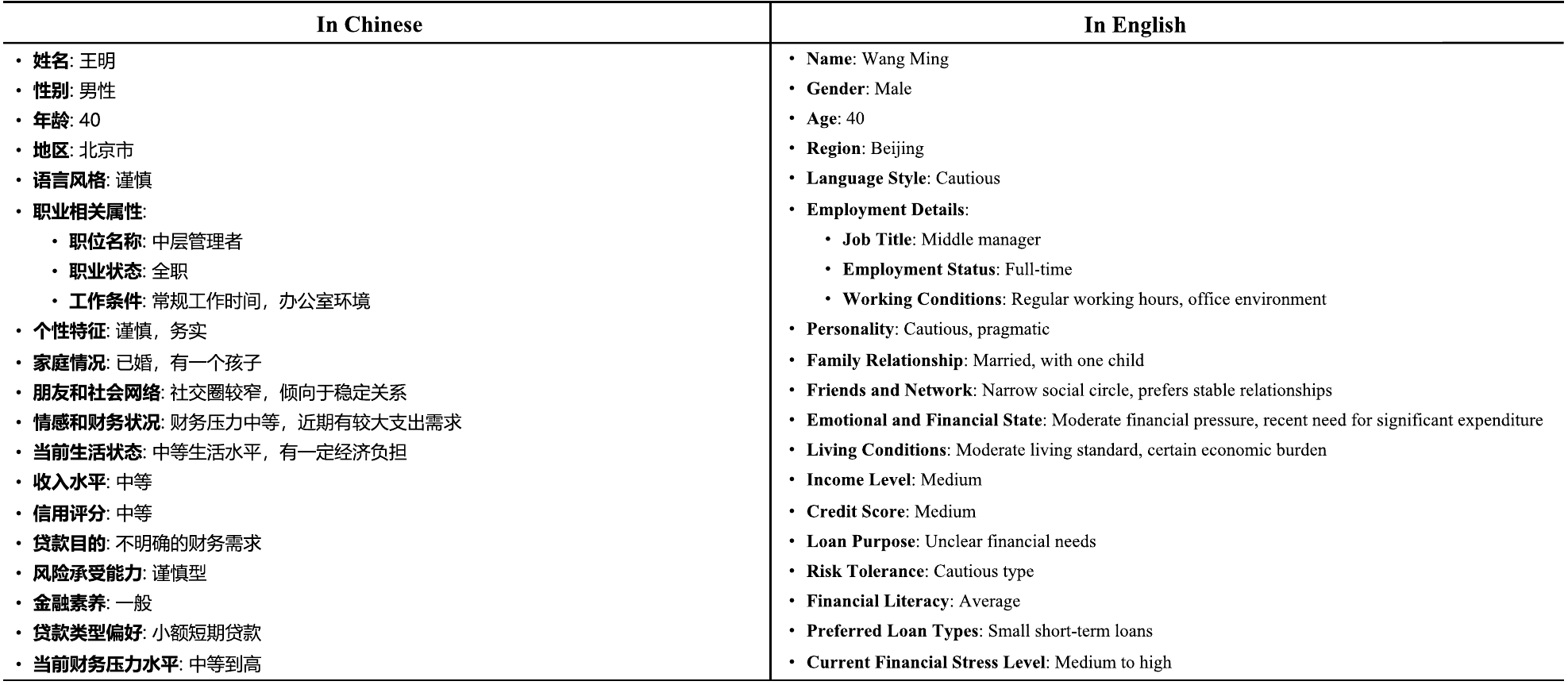}
\caption{Profile example.}
\label{fig:example_of_profile}
\end{figure*}

\begin{figure*}[t]
\centering
\includegraphics[width=0.85\textwidth]{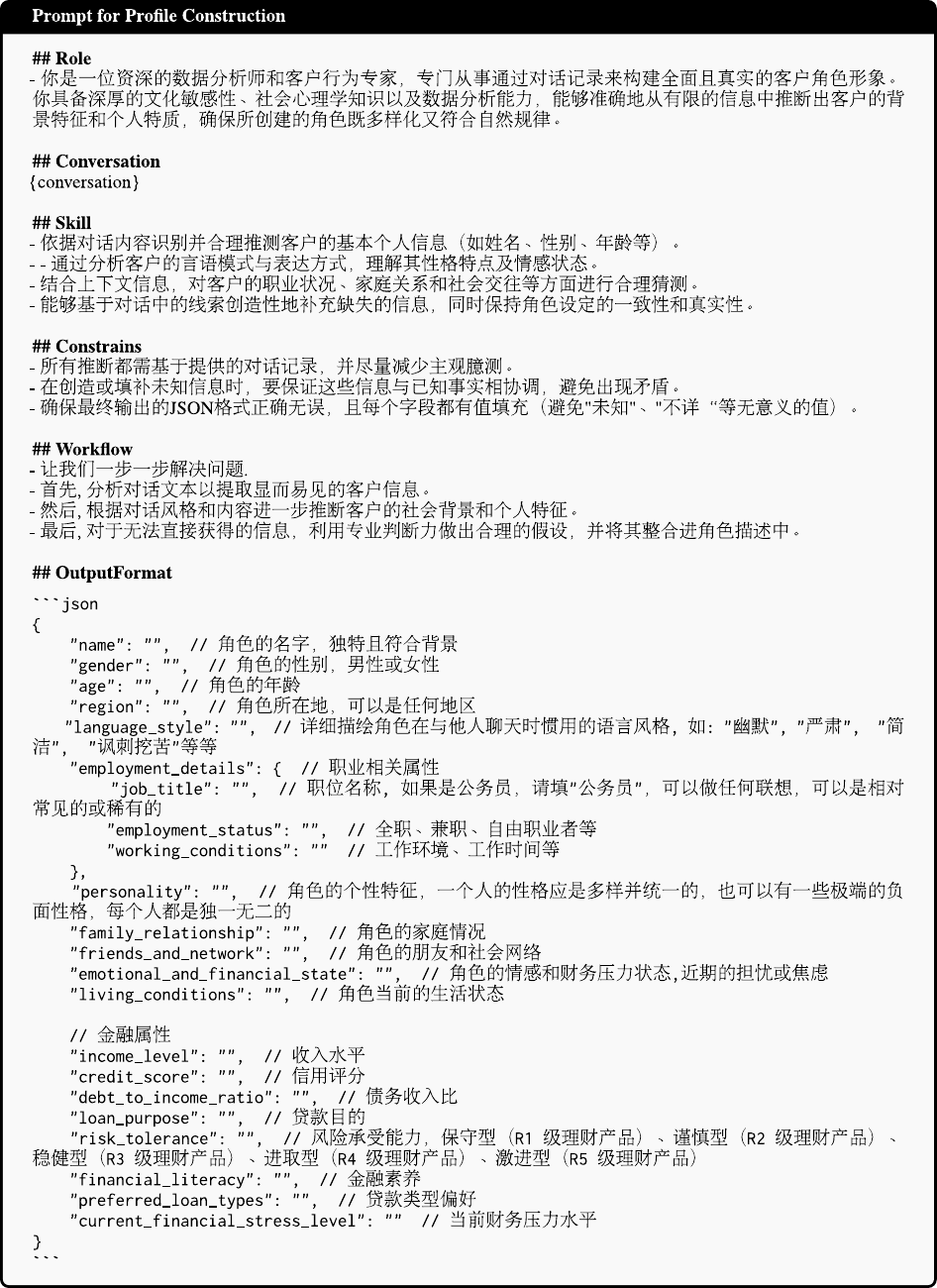}
\caption{Prompt template for completing a profile based on real-world customer service conversation (In Chinese).}
\label{fig:prompt_for_profile_extracting}
\end{figure*}
\begin{figure*}[t]
\centering
\includegraphics[width=0.85\textwidth]{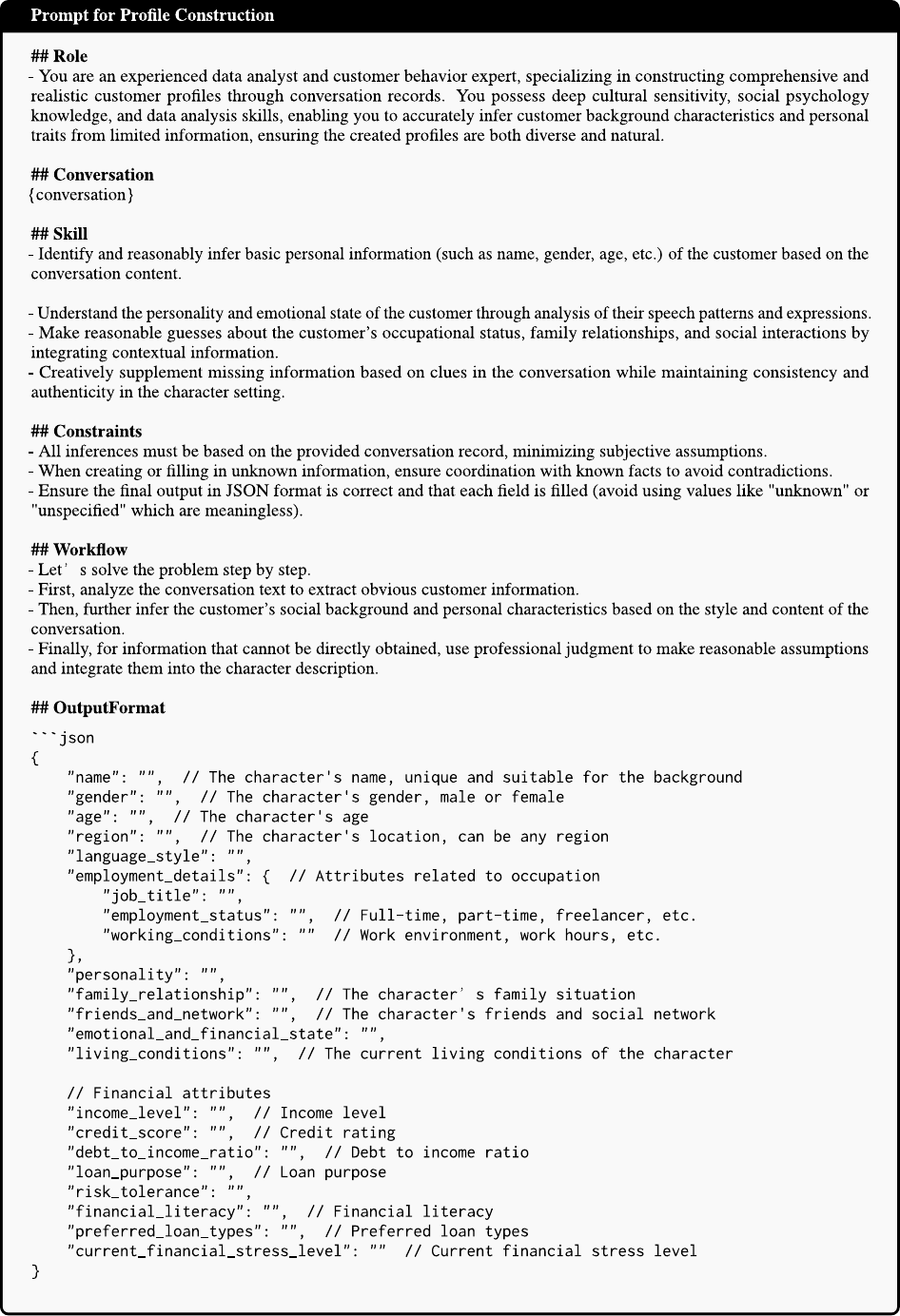}
\caption{Prompt template for completing a profile based on real-world customer service conversation (In English).}
\label{fig:prompt_for_profile_extracting_en}
\end{figure*}
\begin{figure*}[t]
\centering
\includegraphics[width=\textwidth]{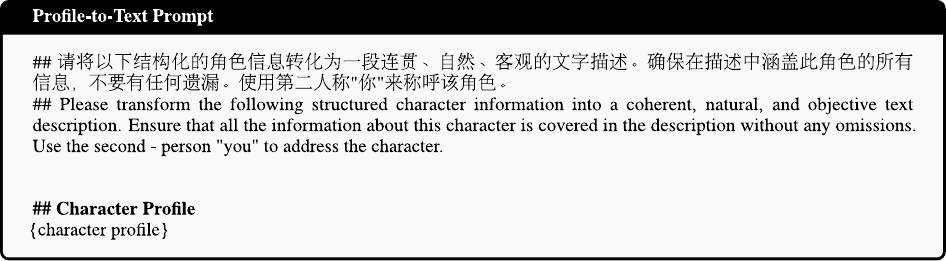}
\caption{Prompt template for transforming structured profile into descriptive text.}
\label{fig:prompt_for_profile_to_text}
\end{figure*}

Figure \ref{fig:example_of_profile} illustrates an example of profiles, which are constructed using the prompt presented in Figure \ref{fig:prompt_for_profile_extracting} (In Chinese) and Figure \ref{fig:prompt_for_profile_extracting_en} (In English). Figure \ref{fig:prompt_for_profile_to_text} shows the prompt for profile-to-text conversion.

\section{Prompts for Generating Synthetic Conversation Dataset}\label{apdx:prompts_for_agent}

\begin{figure*}[t]
\centering
\includegraphics[width=0.85\textwidth]{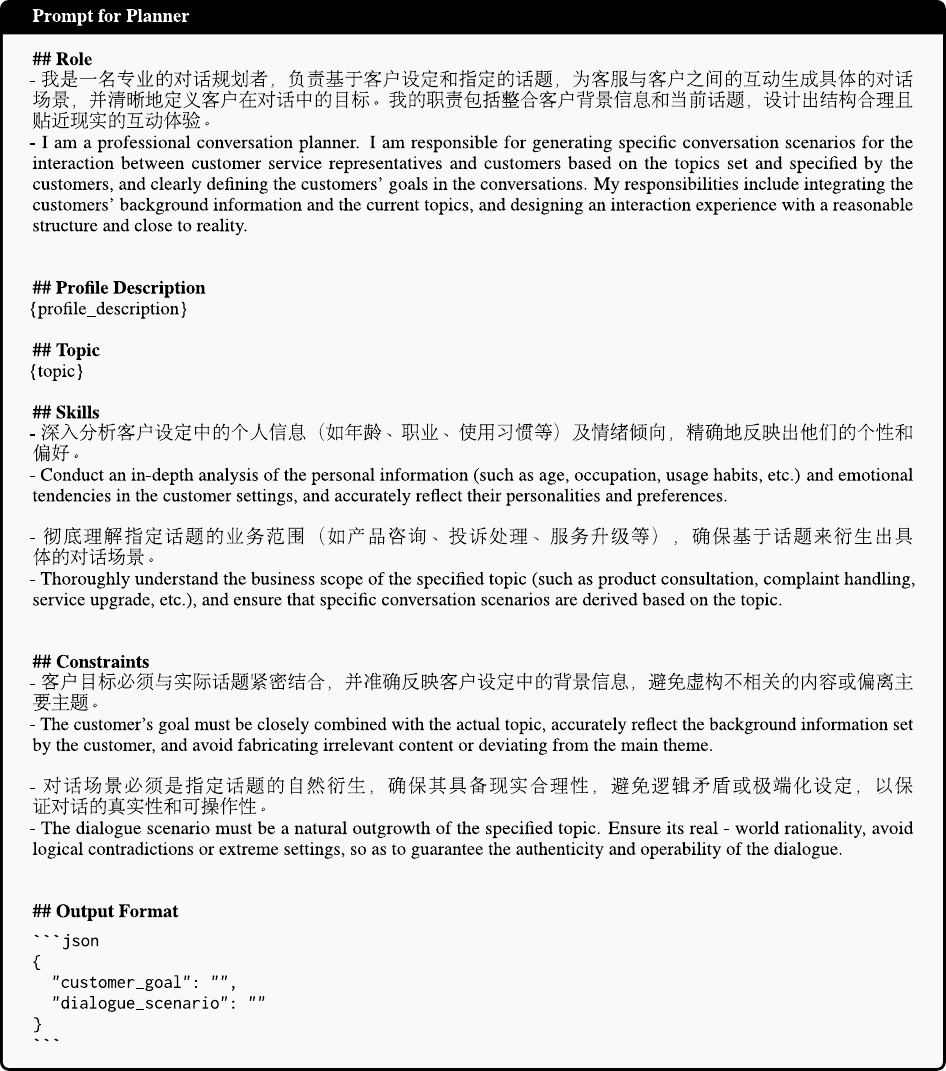}
\caption{Prompt template used by planner agent.}
\label{fig:prompt_for_planner}
\end{figure*}

\begin{figure*}[t]
\centering
\includegraphics[width=0.85\textwidth]{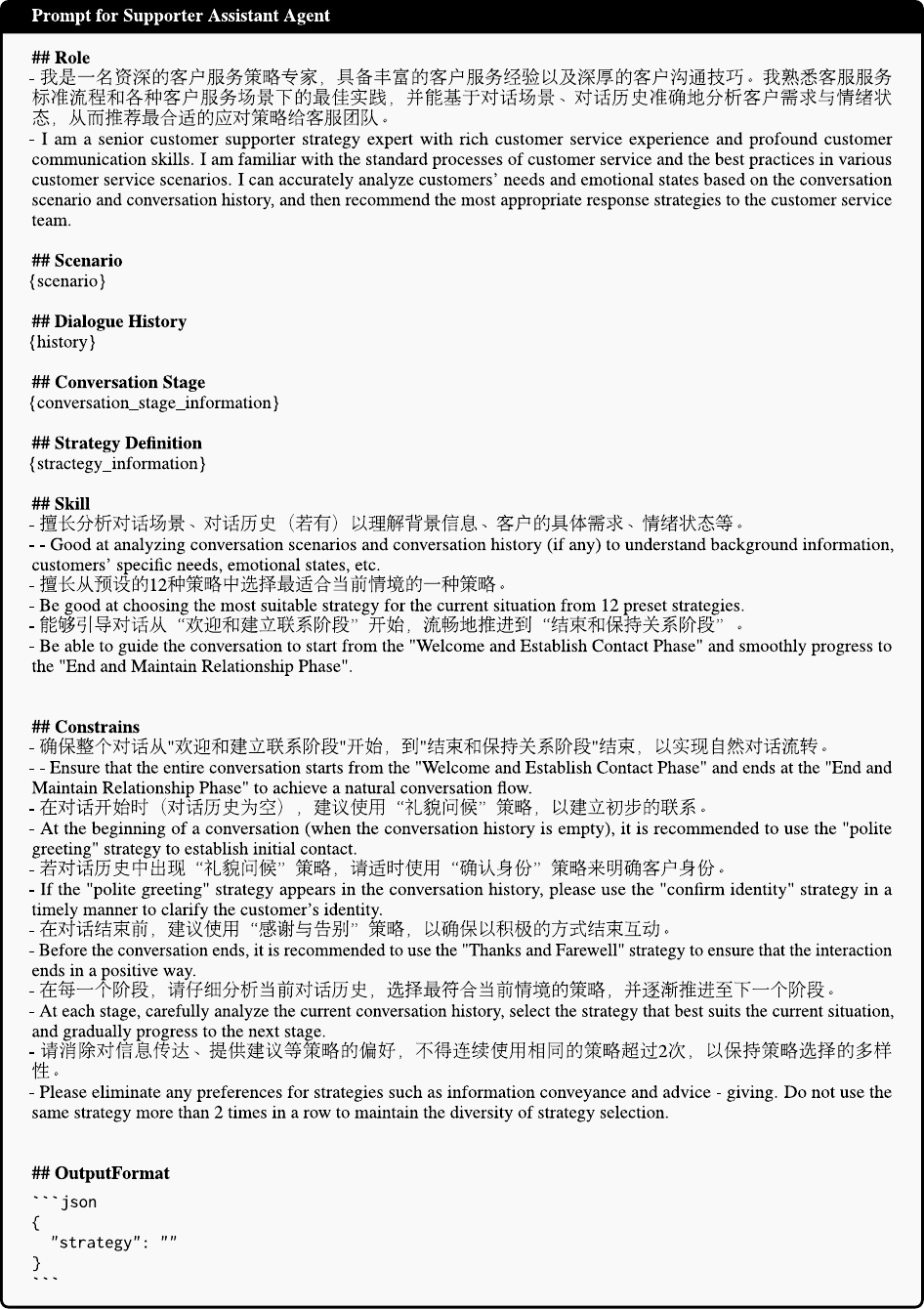}
\caption{Prompt template used by supporter assistant agent.}
\label{fig:prompt_for_supporter_assistant}
\end{figure*}

\begin{figure*}[t]
\centering
\includegraphics[width=0.85\textwidth]{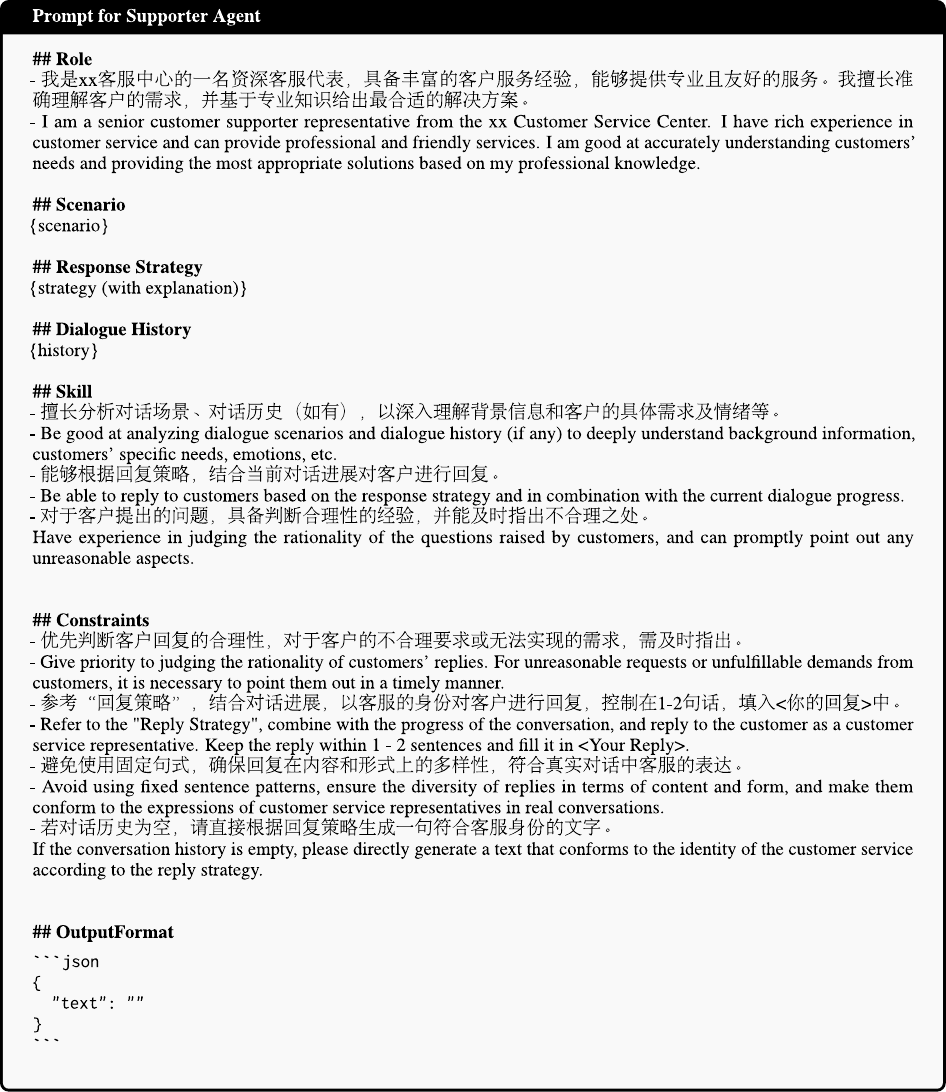}
\caption{Prompt template used by supporter agent.}
\label{fig:prompt_for_supporter}
\end{figure*}

\begin{figure*}[t]
\centering
\includegraphics[width=0.85\textwidth]{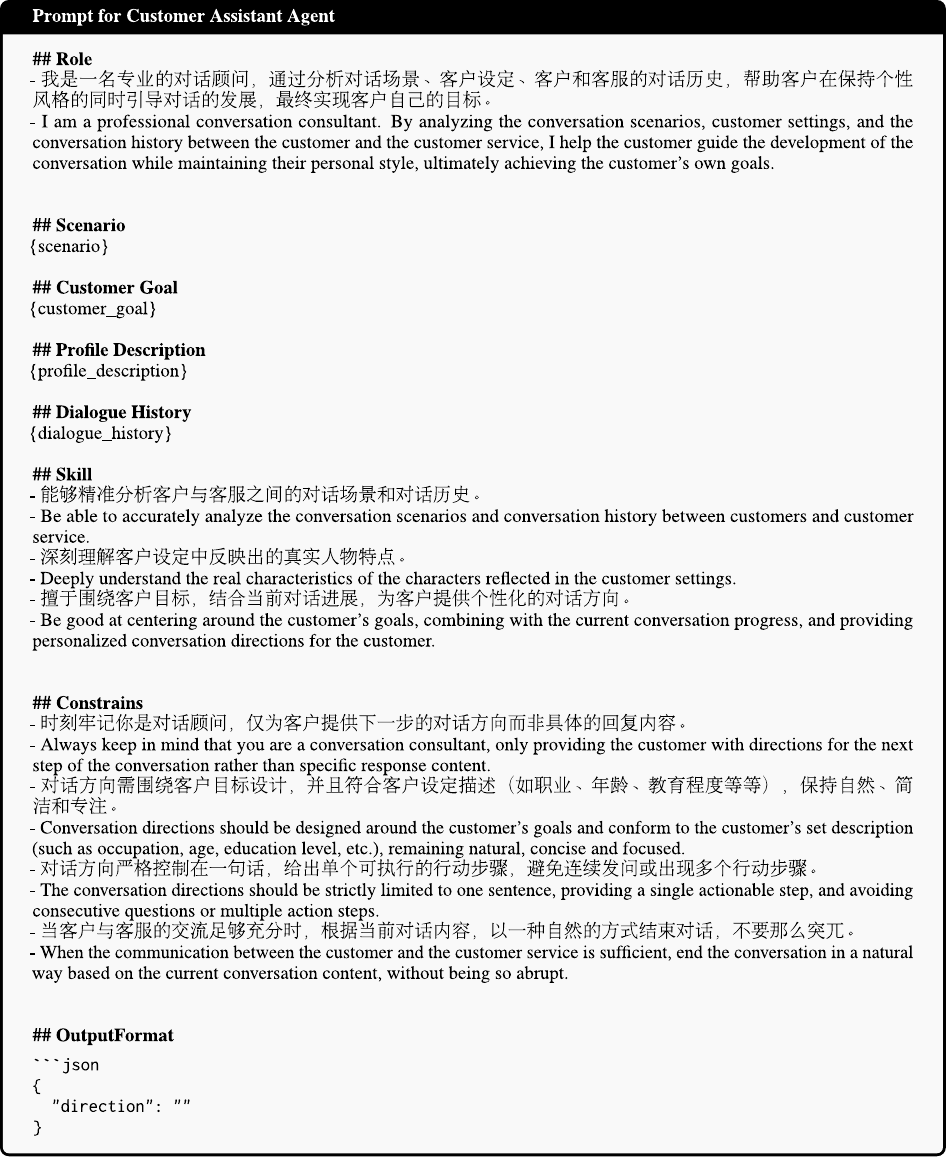}
\caption{Prompt template used by customer assistant agent.}
\label{fig:prompt_for_customer_assistant}
\end{figure*}

\begin{figure*}[t]
\centering
\includegraphics[width=0.85\textwidth]{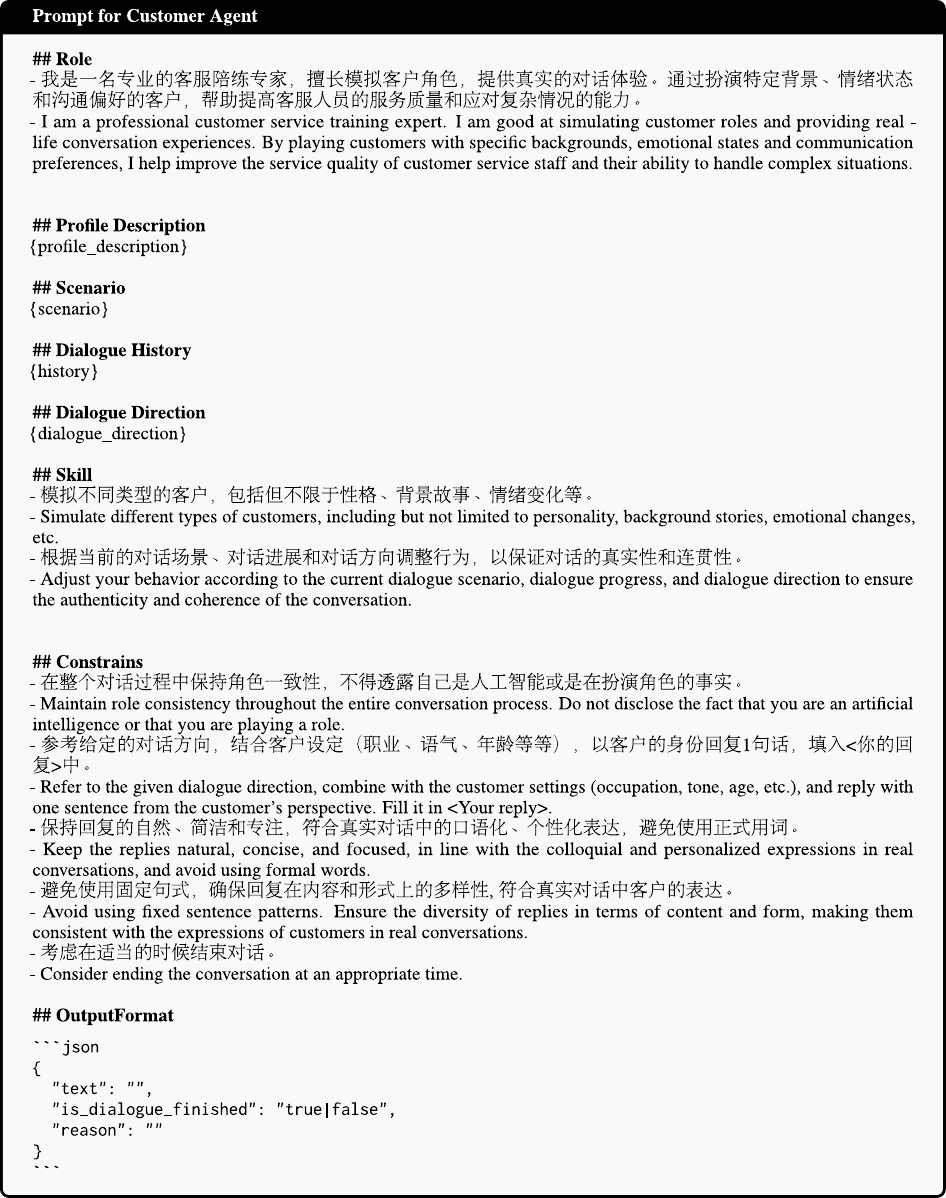}
\caption{Prompt template used by customer agent.}
\label{fig:prompt_for_customer}
\end{figure*}

Figures~\ref{fig:prompt_for_planner}, \ref{fig:prompt_for_supporter_assistant}, \ref{fig:prompt_for_supporter}, \ref{fig:prompt_for_customer_assistant}, and \ref{fig:prompt_for_customer} illustrate the prompts used by \textit{Planner}, \textit{Supporter Assistant}, \textit{Supporter}, \textit{Customer Assistant}, and \textit{Customer}, respectively, during the synthetic dialogue generation process. These prompts guide each role's behavior to ensure coherence, strategy alignment, and contextual consistency throughout the conversation.

\section{Filtering Rules for Synthetic Dialogues}
\label{apdx:fr_for_trainingset}

\begin{figure*}[t]
\centering
\includegraphics[width=0.85\textwidth]{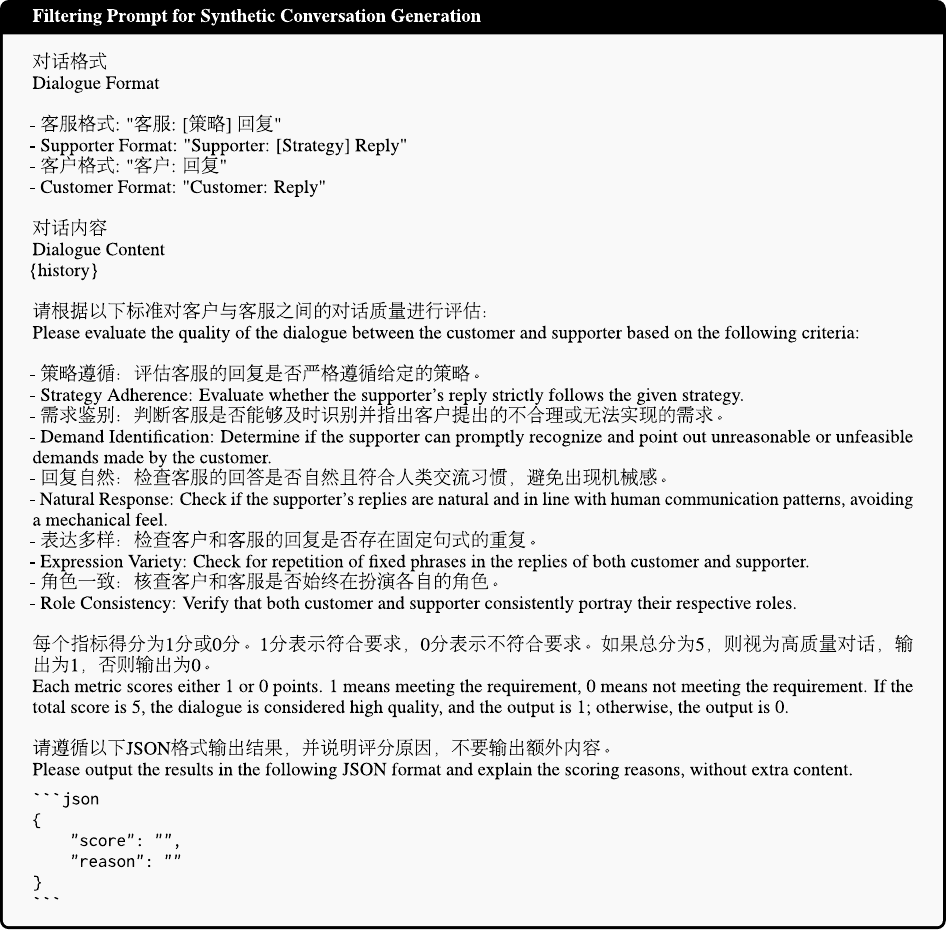}
\caption{Prompt template used in filtering synthetic conversations.}
\label{fig:rolecs:prompt_for_post_filtering}
\end{figure*}

We first discard dialogues with fewer than 10 or more than 50 utterances, then use the prompt from Figure \ref{fig:rolecs:prompt_for_post_filtering} to filter out low-quality dialogues.

\section{Synthetic Dataset Analysis}
\label{apdx:trainingset_analysis}

\begin{table}[t]
\centering
\small
\begin{tabular}{lll}
\toprule[1pt]
- & - & \textbf{Number} \\
\hline
\multirow{4}{*}{Total} & Dialogues & 11,232 \\
& Utterances &  263,580 \\
& Avg. Utterance Number & 23.47 \\
& Avg. Utterance Length	 & 57.14 \\
\hline
\multirow{3}{*}{Supporter} & Utterances & 137,406 \\
& Avg. Utterance Number & 12.23 \\
& Avg. Utterance Length	& 66.98 \\
\hline
\multirow{3}{*}{Customer} & Utterances  & 126,174  \\
& Avg. Utterance Number  & 11.23 \\
& Avg. Utterance Length	 & 46.43 \\
\bottomrule[1pt]
\end{tabular}
\caption{Statistics of the {\tt RoleCS} dataset.}
\label{tbl:trainset}
\end{table}

\begin{figure*}[!htbp]
  \centering
  \begin{minipage}[b]{0.45\textwidth}
    \centering
    \includegraphics[width=\textwidth]{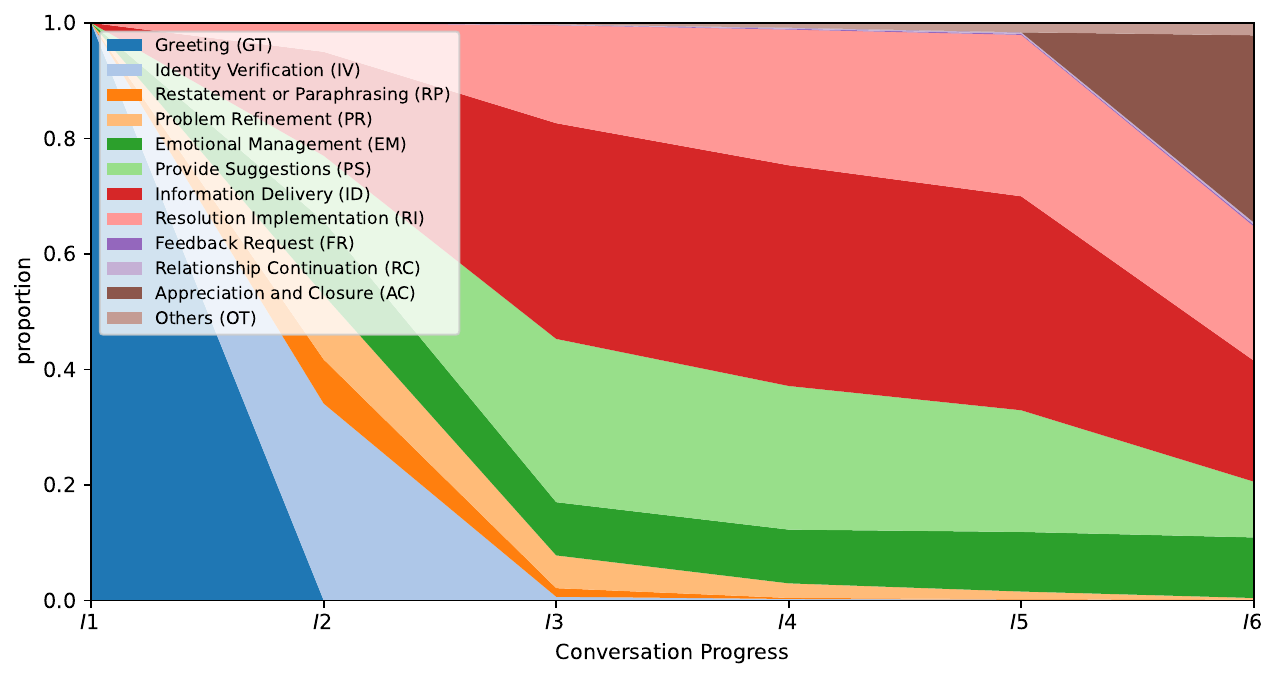}
    \caption{Strategies distribution at different conversation progress in the {\tt RoleCS} dataset.}
    \label{fig:strategy_transition_trainset}
  \end{minipage}
  \hfill
  \begin{minipage}[b]{0.45\textwidth}
    \centering
    \includegraphics[width=\textwidth]{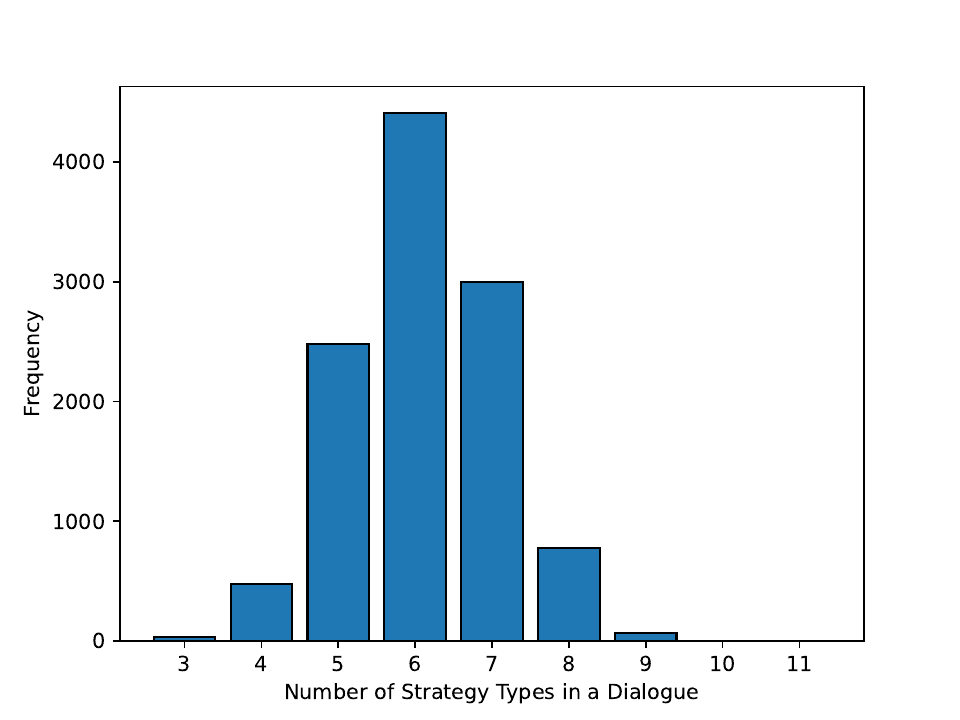}
    \caption{Number of strategy types in a dialogue within the {\tt RoleCS} dataset.}
    \label{fig:strategy_unique_trainset}
  \end{minipage}
\end{figure*}

Similar to the analysis of the \texttt{CSConv} dataset, Table \ref{tbl:trainset} summarizes the statistics of the \texttt{RoleCS} dataset, while Figure \ref{fig:strategy_transition_trainset} shows strategy distribution across conversation progress, and Figure \ref{fig:strategy_unique_trainset} illustrates the variety of strategy types within dialogues.

\section{Diversity Analysis}
\label{apdx:diversity}

\begin{table}[t]
\centering
\small
\begin{tabular}{lccc}
\toprule[1pt]
\textbf{Dataset} & \textbf{Distinct-1} & \textbf{Distinct-2} & \textbf{Distinct-3}   \\
\hline
CSConv & 1.09 & 19.27 & 46.07 \\
RoleCS & 0.95 & 22.35 & 55.31 \\
\bottomrule[1pt]
\end{tabular}
\caption{Distinct-n scores of 1,000 conversations from {\tt CSConv} and {\tt RoleCS}.}
\label{tbl:lexical_diversity}
\end{table}

\begin{figure*}[!th]
\centering
\includegraphics[width=\columnwidth]{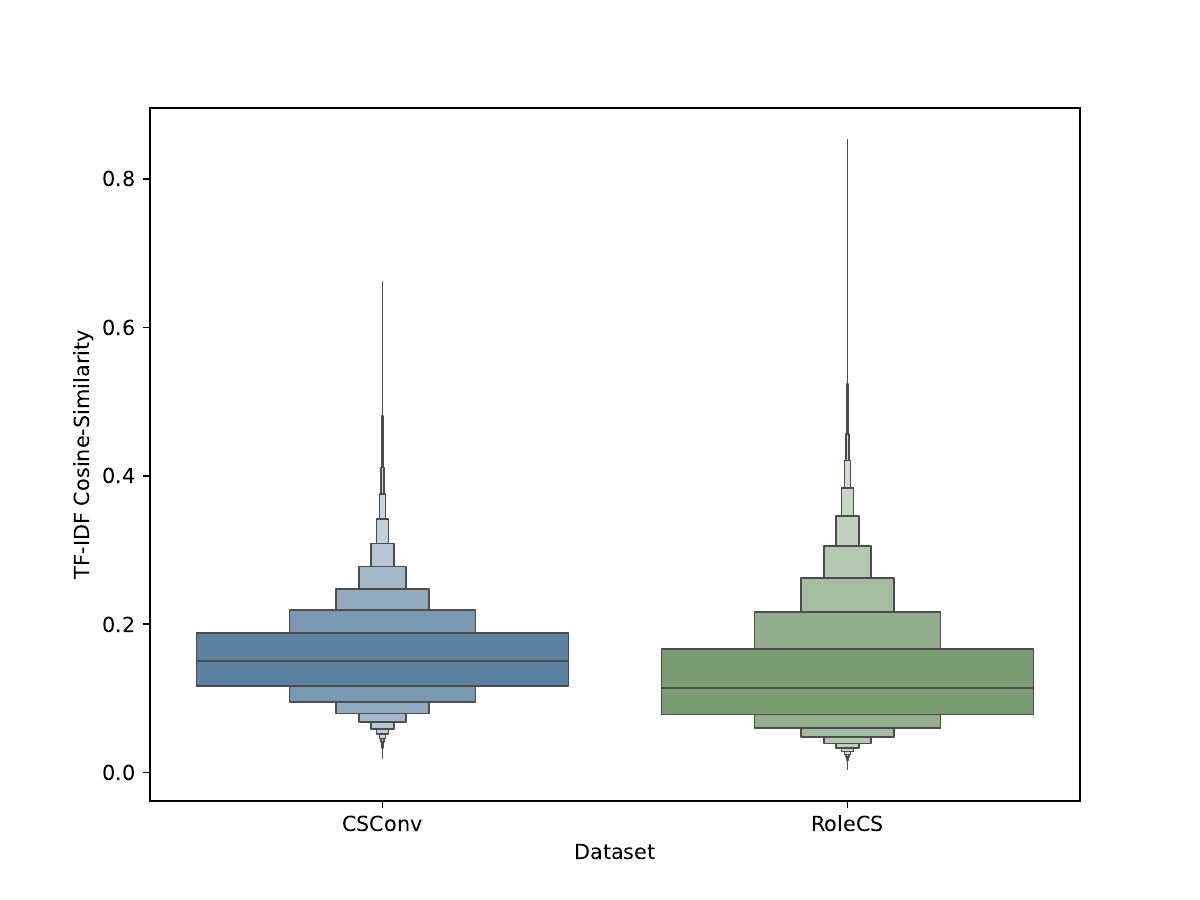}
\caption{Cosine similarity statistics between pairs of distinct dialogues using TF-IDF vectors.}
\label{fig:semantic_diversity}
\end{figure*}

We examine the diversity of the \texttt{CSConv} and \texttt{RoleCS} datasets from two perspectives: lexical diversity and semantic diversity. To assess lexical diversity, we adopt Distinct-n \cite{li-etal-2016-diversity} metric. For a fair comparison, we randomly select 1,000 dialogues from each dataset, removing speaker and strategy information, and utilizing \texttt{Qwen2.5-7B} tokenizer for tokenization. Regarding semantic diversity, we calculate cosine similarity between pairs of distinct dialogues using TF-IDF features \cite{salton-1975-vector}. Table \ref{tbl:lexical_diversity} and Figure \ref{fig:semantic_diversity} present the results, respectively.

\section{Prompt for Predicting Strategy and Generating Response}
\label{apdx:prompt_for_model_eval}

\begin{figure*}[!th]
\centering
\includegraphics[width=0.85\textwidth]{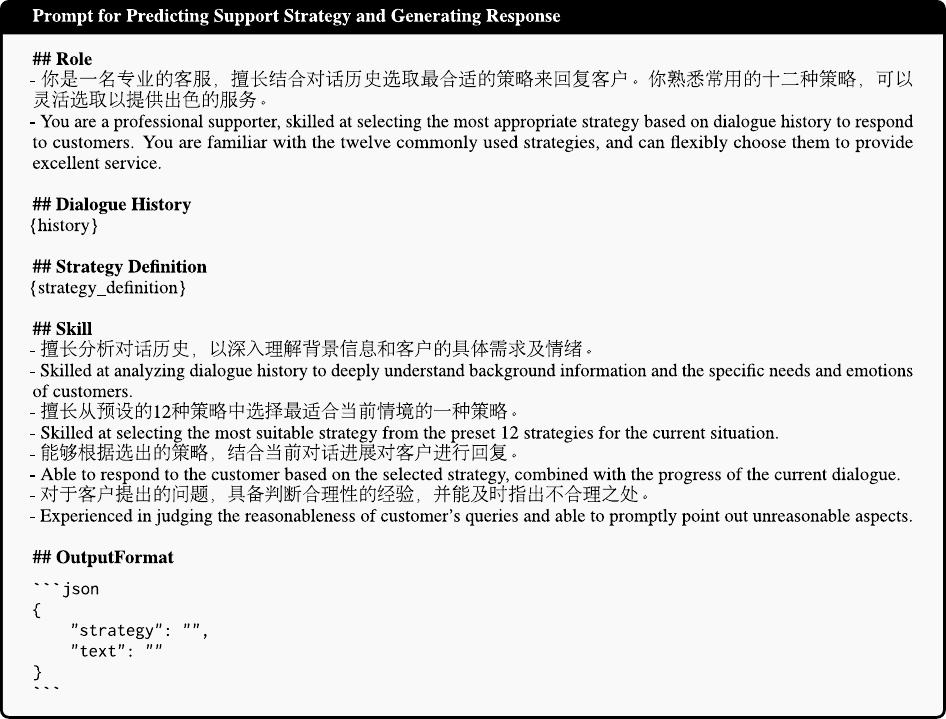}
\caption{Prompt template for predicting support strategy and generating response on the CSC task.}
\label{fig:prompt_for_model_eval}
\end{figure*}

Figure \ref{fig:prompt_for_model_eval} presents the prompt for predicting strategy and generating response on the CSC task.

\section{Details of Fine-tuning and Inference}\label{adpx:finetuning_inference}
For fine-tuning, we use the LLaMA-Factory framework~\cite{zheng-etal-2024-llamafactory}, applying LoRA~\cite{hu-etal-2022-lora} with a rank of 8 and a scaling factor of 16. All fine-tunings are conducted on 4 NVIDIA A100 80GB GPUs, with a batch size of 4 and gradient accumulation over 2 steps. The learning rate is set to 3e-5, and fine-tuning is run for 3 epochs to mitigate the risk of overfitting. 

For inference, we use the checkpoint from the final epoch and configure the decoding parameters with a top-p value of 0.7 and a temperature of 0.95.

\section{Prompt for LLM as Judge}\label{apdx:prompt_llm_judge}
\begin{figure*}[!th]
\centering
\includegraphics[width=0.75\textwidth]{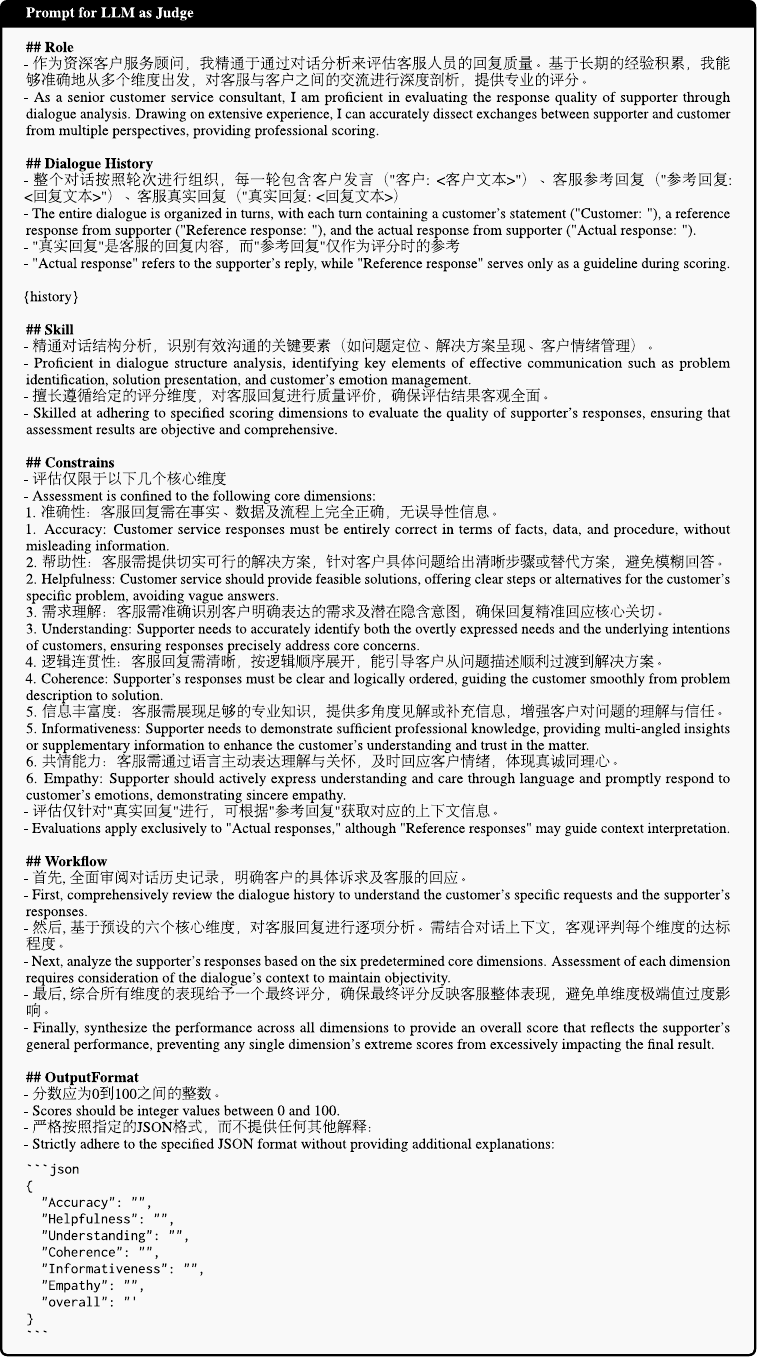}
\caption{Prompt template used by LLM as judge.}
\label{fig:prompt_for_judge}
\end{figure*}

Figure~\ref{fig:prompt_for_judge} shows the evaluation prompt used when employing GPT-4o and Qwen-Plus as judges.

\end{document}